\documentclass{article} 

\usepackage{natbib}

\usepackage{iclr2017_arXiv,times}

\usepackage{hyperref}
\usepackage[autostyle]{csquotes}
\usepackage[textsize=tiny,disable]{todonotes}
\usepackage[utf8]{inputenc}

\title{Reinforcement Learning in Practice: \\Opportunities and Challenges}

\author{Yuxi Li (yuxili@gmail.com)}
\date{\vspace{-5ex}}

\begin{document}

\maketitle








\begin{abstract} 

This article is a gentle discussion about the field of reinforcement learning in practice, about opportunities and challenges, touching a broad range of topics, with perspectives and without technical details. 
The article is based on both historical and recent research papers, surveys, tutorials, talks, blogs, books, (panel) discussions, and workshops/conferences. 
Various groups of readers, like researchers, engineers, students, managers,  investors, officers, and people wanting to know more about the field, may find the article interesting.

In this article, we first give a brief introduction to reinforcement learning (RL), and its relationship with deep learning, machine learning and AI.
Then we discuss opportunities of RL, in particular, 
products and services, games, bandits, recommender systems, robotics, transportation, finance and economics, healthcare, education, combinatorial optimization, computer systems, and science and engineering.
Then we discuss challenges, in particular,
1) foundation, 2) representation, 3) reward, 4) exploration, 5) model, simulation, planning, and benchmarks, 
6) off-policy/offline learning, 7) learning to learn a.k.a. meta-learning, 8) explainability and interpretability, 9) constraints, 10) software development and deployment, 11) business perspectives, and 12) more challenges.
We conclude with a discussion, attempting to answer: \enquote{Why has RL not been widely adopted in practice yet?} and \enquote{When is RL helpful?}.

\end{abstract}  

\newpage
\tableofcontents
\newpage

\section{Introduction}
\label{}

There are more and more exciting AI achievements like AlphaGo~\citep{Silver-AlphaGo-2016, Littman2021}.
Deep learning and reinforcement learning are underlying techniques. 
Besides games, reinforcement learning has been making tremendous progress in diverse areas like recommender systems and robotics. These successes have sparked rapidly growing interests in using reinforcement learning to solve many other real life problems. 

Reinforcement learning (RL) refers to the general problem of learning a behavior that optimizes a long-term performance metric in a sequential setting. RL techniques can be used to tackle goal-directed or optimization problems that can be transformed into sequential decision making problems. 
As such, RL overlaps largely with optimal control and operations research, with strong ties to optimization, statistics, game theory, causal inference, sequential experimentation, etc., and applicable broadly to many problems in science, engineering and arts~\citep{Bertsekas2019, Littman2015, Powell2021, Sutton2018, Szepesvari2010}.



With recent exciting achievements in deep learning~\citep{LeCun2015, Bengio2021, Dean2022deep}, benefiting from big data, powerful computation, new algorithmic techniques, mature software packages and architectures, and strong financial support, we have been witnessing the renaissance of RL~\citep{Krakovsky2016}, especially the combination of DL and RL, i.e., deep RL~\citep{Li2017DeepRL}. 
The integration of RL and neural networks has a long history, see e.g., \citet{Sutton2018,Bertsekas96, Schmidhuber2015-DL}, with a prominent example,  Backgammon \citep{Tesauro1994}. 

\cite{Silver2020} presents the following conjectures. Reinforcement learning defines the objective. Conjecture 1: RL is enough to formalise the problem of intelligence. Deep learning (DL) gives mechanisms for optimising objectives. Conjecture 2: Deep neural networks can represent and learn any computable function. Deep RL combines the RL problem with DL solution. Conjecture 3: RL + DL can solve the problem of intelligence. 
\cite{Silver2021} present the  Reward-is-Enough hypothesis: \enquote{Intelligence, and its associated abilities, can be understood as subserving the maximisation of reward by an agent acting in its environment.}

Russell and Norvig's AI textbook states that \enquote{reinforcement learning might be considered to encompass all of AI: an agent is placed in an environment and must learn to behave successfully therein} and \enquote{reinforcement learning can be viewed as a microcosm for the entire AI problem}~\citep{Russell2009}. It is also shown that tasks with computable descriptions in computer science can be formulated as RL problems~\citep{Hutter2005}.

It is desirable to have RL systems that work in the real world with real benefits. However, there are many theoretical, algorithmic, and practical challenges before RL is widely applied: generalization, sample efficiency, exploration vs. exploitation dilemma, credit assignment, safety, explainability, and technical debts, just name a few.

To share knowledge and lessons and to identify key research challenges, 
for the topic of RL for real life,
we organized workshops in ICML 2019 and ICML 2021, as well as a virtual workshop in 2020.
We brought together researchers and practitioners from industry and academia interested in addressing practical and/or theoretical issues that arise when applying RL to real life scenarios.  
Researchers presented some of their latest results, shared first-hand lessons and experience from real life deployments, and identified key research problems and open challenges.
We also guest-edited a special issue for the Machine Learning journal. 
More information can be found on the website: \url{https://sites.google.com/view/RL4RealLife}. 


This article is a gentle discussion about the field of reinforcement learning for real life, about opportunities and challenges, touching a broad range of topics, with perspectives and without technical details. 
The article is based on both historical and recent research papers, surveys, tutorials, talks, blogs, books, (panel) discussions, and workshops/conferences. 
Various groups of readers, like researchers, engineers, students, managers,  investors, officers, and people wanting to know more about the field, may find the article interesting.
This article is not a comprehensive overview.
As a result, it definitely misses citing many important and excellent papers.
There will inevitably be shortcomings and even errors, when writing an article covering so many topics.
Comments and criticisms are always welcome. 

We organize the rest of the article as follows.
We first give a brief introduction to RL, and its relationship with deep learning, machine learning and AI.
Then we discuss opportunities of RL, in particular, 
products and services, games, bandits, recommender systems, robotics, transportation,  finance and economics, healthcare, education, combinatorial optimization, computer systems, and science and engineering.
Then we discuss challenges, in particular,
1) foundation, 2) representation, 3) reward, 4) exploration, 5) model, simulation, planning, and benchmarks, 
6) off-policy/offline learning, 7) learning to learn a.k.a. meta-learning, 8) explainability and interpretability, 9) constraints, 10) software development and deployment, 11) business perspectives, and 12) more challenges.
We conclude with a discussion, attempting to answer: \enquote{Why has RL not been widely adopted in practice yet?} and \enquote{When is RL helpful?}.

\section{A Brief Introduction to RL}

\begin{figure}[h]
\includegraphics[width=1.0\linewidth]{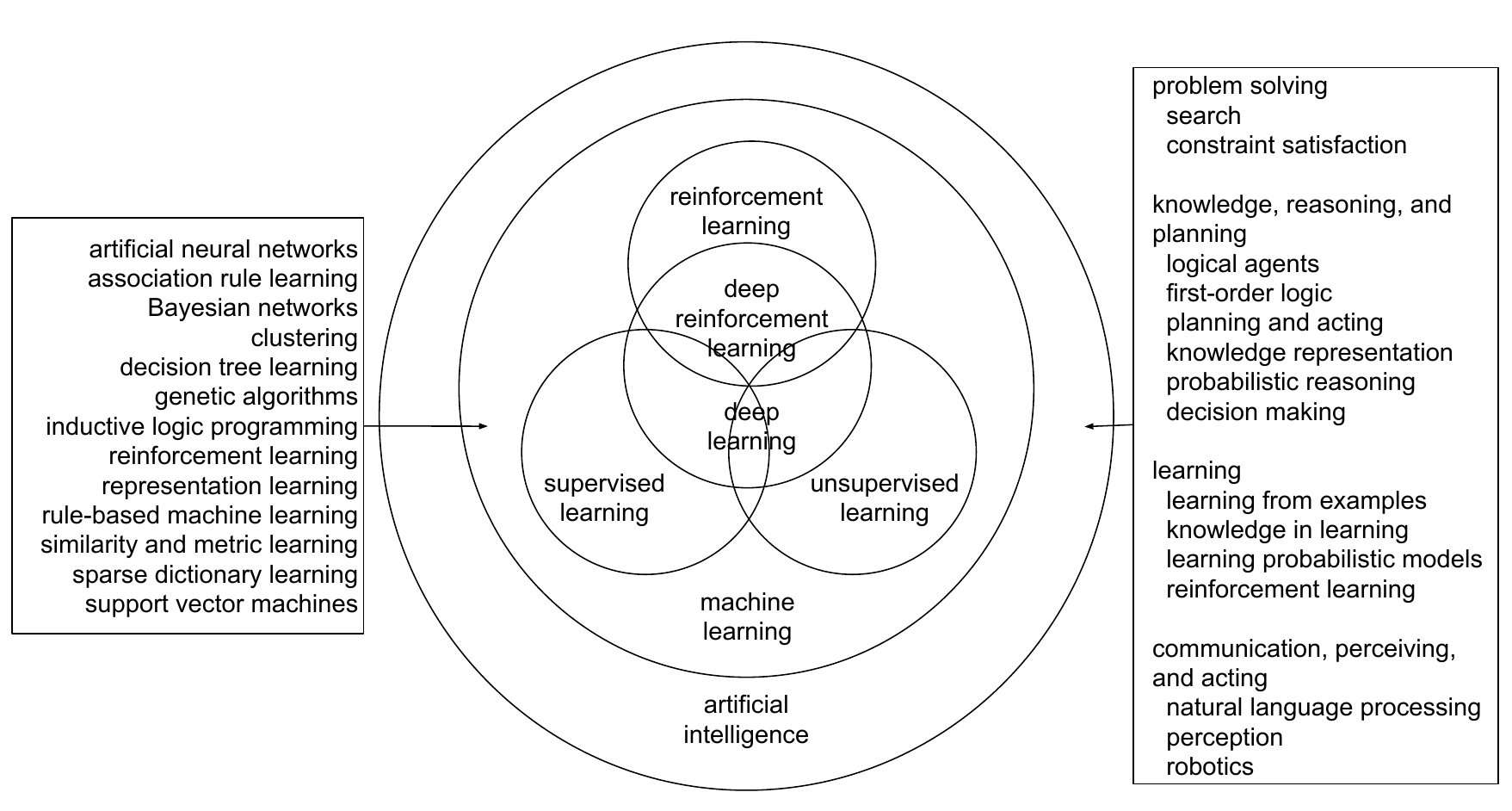}
\caption{Relationship among reinforcement learning,  deep learning, deep reinforcement learning, supervised learning,  unsupervised learning,  machine learning, and AI. Deep reinforcement learning, as the name indicates, is at the intersection of deep learning and reinforcement learning.
We usually categorize machine learning as supervised learning, unsupervised learning, and reinforcement learning.
However, they may overlap with each other.
Deep learning can work with/as these and other machine learning approaches. 
Deep learning is part of machine learning, which is part of AI.
The approaches in machine learning on the left is based on Wikipedia, \url{https://en.wikipedia.org/wiki/Machine_learning}.
The approaches in AI on the right is based on~\cite{Russell2009}.
Note that all these fields are evolving,
e.g., deep learning and reinforcement learning are addressing many classical AI problems, 
such as logic, reasoning, and knowledge representation.
}
\label{relation}
\end{figure}

Machine learning is about learning from data and making predictions and/or decisions.
We usually categorize machine learning into supervised learning, unsupervised learning, and reinforcement learning. 
Supervised learning works with labeled data,
including classification and regression with categorical and numerical outputs, respectively.
Unsupervised learning finds patterns from unlabelled data, e.g., clustering, principle component analysis (PCA) and generative adversarial networks (GANs).
In RL, there are evaluative feedbacks but no supervised labels. 
Evaluative feedbacks can not indicate whether a decision is correct or not, as labels in supervised learning.
Supervised learning is usually one-step, and considers only immediate cost or reward, whereas RL is sequential, and ideal RL agents are far-sighted and consider long-term accumulative rewards.
RL has additional challenges like credit assignment and exploration vs. exploitation, comparing with supervised learning.
Moreover, in RL, an action can affect next and future states and actions, which results in distribution shift inherently.
Deep learning (DL), or deep neural networks (DNNs), can work with/as these and other machine learning approaches. 
Deep learning is part of machine learning, which is part of AI. 
Deep RL is an integration of deep learning and RL.\footnote{RL can be viewed as equivalent to AI, as discussed earlier.
As a framework for decision making, RL is beyond prediction, which is usually the task by ML, usually refering to supervised learning and unsupervised learning. As a result, the categorization may appear confusing and contradictive. Here we present both perspectives: 1) that DL and RL are part of ML, which is part of AI, and 2) that RL is equivalent to AI.} 
Figure~\ref{relation} presents the relationship among these concepts.



\begin{figure}
\centering
\includegraphics[width=0.5\linewidth]{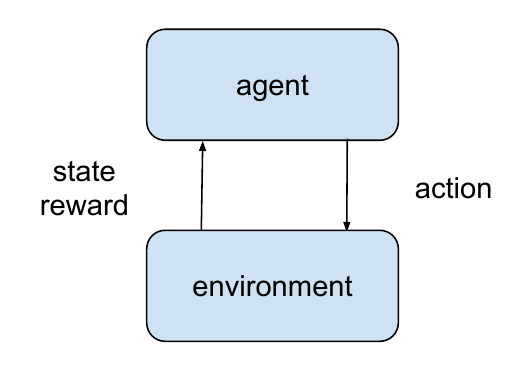}
\caption{Agent-environment interaction.}
\label{agent}
\end{figure}

An RL agent interacts with the environment over time to learn a policy, by trial and error, that maximizes the long-term, cumulated reward. At each time step, the agent receives an observation, selects an action to be executed in the environment, following a policy, which is the agent's behaviour, i.e., a mapping from an observation to actions. The environment responds with a scalar reward and by transitioning to a new state according to the environment dynamics. 
Figure~\ref{agent} illustrates the agent-environment interaction.

In an episodic environment, this process continues until the agent reaches a terminal state and then it restarts.
Otherwise, the environment is continuing without a terminal state.
There is a discount factor to measure the influence of future award.
The model refers to the transition probability and the reward function. 
The RL formulation is very general:
state and action spaces can be discrete or continuous;
an environment can be a multi-armed bandit, an MDP, a partially observable MDP (POMDP), a game, etc.;
and an RL problem can be deterministic, stochastic, dynamic, or adversarial.

A state or action value function measures the goodness of each state or state action pair, respectively.
It is a prediction of the return, or the expected, accumulative, discounted, future reward.
The action value function is usually called the $Q$ function.
An optimal value is the best value achievable by any policy, and the corresponding policy is an optimal policy.
An optimal value function encodes global optimal information, 
i.e., it is not hard to find an optimal policy based on an optimal state value function, 
and it is straightforward to find an optimal policy with an optimal action value function.  
The agent aims to maximize the expectation of a long-term return or to find an optimal policy.

When the system model is available, we may use dynamic programming methods: 
policy evaluation to calculate value/action value function for a policy, 
and value iteration or policy iteration for finding an optimal policy, 
where policy iteration consists of policy evaluation and policy improvement. 
Monte Carlo (MC) methods learn from complete episodes of experience, not assuming knowledge of transition nor reward models, and use sample means for estimation. 
Monte Carlo methods are applicable only to episodic tasks.
In model-free methods, the agent learns with trail and error from experience directly; 
the model, usually the state transition, is not known. 
RL methods that use models are model-based methods; 
the model may be given or learned from experience.

The prediction problem, or policy evaluation, is to compute the state or action value function for a policy. The control problem is to find the optimal policy. Planning constructs a value function or a policy with a model.

In an online mode, an algorithm is trained on data available in sequence. 
In an offline mode, or a batch mode, an algorithm is trained on a collection of data.

Temporal difference (TD) learning learns state value function directly from experience, with bootstrapping from its own estimation, in a model-free, online, and fully incremental way.
With bootstrapping, an estimate of state or action value is updated from subsequent estimates.
TD learning is an on-policy method, with samples from the same target policy.
 Q-learning is a temporal difference control method, learning an optimal action value function to find an optimal policy.
Q-learning is an off-policy method, learning with experience trajectories from some behaviour policy, but not necessarily from the target policy. 
The notion of on-policy and off-policy can be understood as \enquote{same-policy} and \enquote{different-policy}, respectively.

In tabular cases, a value function and a policy are stored in  tabular forms. 
Function approximation is a way for generalization when the state and/or action spaces are large or continuous. 
Function approximation aims to generalize from examples of a function to construct an approximate of the entire function.
Linear function approximation is a popular choice, partially due to its desirable theoretical properties. 
A function is approximated by a linear combination of basis functions, which usually need to be designed manually.
The coefficients, or weights, in the linear combination, need to be found by learning algorithms.

We may also have non-linear function approximation, in particular, with DNNs, to represent the state or observation and/or actions, to approximate value function, policy, and model (state transition function and reward function), etc. 
Here, the weights in DNNs need to be found.
We obtain deep RL methods when we integrate deep learning with RL. 
Deep RL is popular and has achieved stunning achievements recently.

TD learning and Q-learning are value-based methods.
In contrast, policy-based methods optimize the policy directly, e.g., policy gradient.
Actor-critic algorithms update both the value function and the policy.

There are several popular deep RL algorithms.
DQN integrates Q-learning with DNNs, and utilizes the experience replay and a target network to stabilize the learning.
In experience replay, experience or observation sequences, i.e., sequences of state, action, reward, and next state, are stored in the replay buffer, and sampled randomly for training.
A target network keeps its separate network parameters, which are for the training, and updates them only periodically, rather than for every training iteration.
\cite{Mnih-A3C-2016} present Asynchronous Advantage Actor-Critic (A3C), in which parallel actors employ different exploration policies to stabilize training, and the experience replay is not utilized.
Deterministic policy gradient can help estimate policy gradients more efficiently.
\cite{Silver-DPG-2014} present Deterministic Policy Gradient (DPG) and \citet{Lillicrap2016}  extend it to Deep DPG (DDPG).
Trust region methods are an approach to stabilize policy optimization by constraining gradient updates.
\cite{Schulman2015} present Trust Region Policy Optimization (TRPO) and 
\cite{Schulman2017PPO} present Proximal Policy Optimization (PPO).
\cite{Haarnoja2018} present Soft Actor-Critic (SAC), an off-policy algorithm aiming to simultaneously succeed at the task and act as randomly as possible.
\cite{Fujimoto2018} present Twin Delayed Deep Deterministic policy gradient algorithm (TD3) to minimize the effects of overestimation on both the actor and the critic.
\cite{Fujimoto2021} present a variant of TD3 for offline RL.

A fundamental dilemma in RL is the exploration vs. exploitation tradeoff.
The agent needs to exploit the currently best action to maximize rewards greedily, yet it has to explore the environment to find better actions, when the policy is not optimal yet, or the system is non-stationary.
A simple exploration approach is $\epsilon$-greedy, 
in which an agent selects a greedy action with probability $1-\epsilon$, and a random action otherwise.

\subsubsection*{A Shortest Path Example}

Consider the shortest path problem as an example. 
The single-source shortest path problem is to find the shortest path between a pair of nodes to minimize the distance of the path, or the sum of weights of edges in the path. 
We can formulate the shortest path problem as an RL problem.
The state is the current node. 
At each node, following the link to each neighbour is an action. 
The transition model indicates that the agent goes to a neighbour after choosing a link to follow. 
The reward is then the negative of link cost. 
The discount factor can be 1, so that we do not differentiate immediate reward and future reward. 
It is fine since it is an episodic problem.
The goal is to find a path to maximize the negative of the total cost, i.e., to minimize the total distance. 
An optimal policy is to choose the best neighbour to traverse to achieve the shortest path; 
and, for each state/node, an optimal value is the negative of the shortest distance from that node to the destination. 

\begin{figure}
\centering
\includegraphics[width=0.75\linewidth]{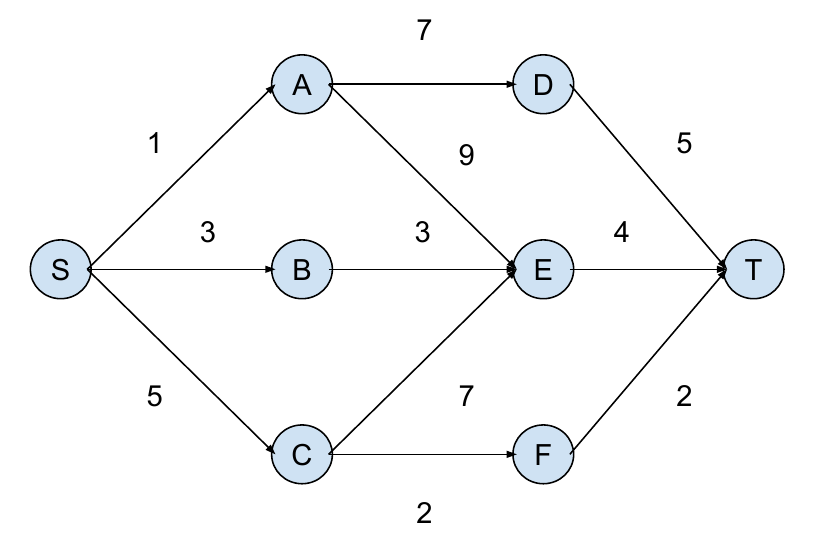}
\caption{An shortest path example.}
\label{shortestpath}
\end{figure}

An example is shown in Figure~\ref{shortestpath}, 
with the graph nodes, (directed) links, and costs on links.
We want to find the shortest path from node $S$ to node $T$.
We can see that if we choose the nearest neighbour of node $S$, i.e., node $A$, as the next node to traverse, then we cannot find the shortest path, i.e., $S \rightarrow C \rightarrow F \rightarrow T$.
This shows that a short-sighted choice, e.g., choosing node $A$ to traverse from node $S$, may lead to a sub-optimal solution.
RL methods, like TD-learning and Q-learning, can find the optimal solution by considering long-term rewards.

Some readers may ask why not use Dijkstra's algorithm.
Dijkstra's algorithm is an efficient algorithm, with the global information of the graph, including nodes, edges, and weights. 
RL can work without such global graph information by wandering in the graph according to some policy, i.e., in a model-free approach. 
With such graph information, Dijkstra's algorithm is efficient for the particular shortest path problem.
However, RL is more general than Dijkstra's algorithm, applicable to many problems, e.g., a problem with stochastic weights or a new instance with meta-learning.

\section{Opportunities}
\label{success}

Some years ago we talked about potential applications of machine learning in industry, and machine learning is already in many large scale production systems. 
Now we are talking about RL, and RL starts to appear in production systems.
Contextual bandits are a \enquote{mature} technique that can be widely applied.
RL is \enquote{mature} for many single-, two-, and multi-player games.
We will probably see more results in recommendation/personalization soon.
System optimization and operations research are expected to draw more attention.
Robotics and healthcare are already at the center of attention.
In the following, we discuss several successful applications.
Note, topics in subsections may overlap, e.g, robotics may also appear in education and healthcare, computer systems are involved in all topics, combinatorial optimization can be applied widely, like in transportation and computer systems, and science and engineering are general areas. 
There are vast number of papers about RL applications.
See e.g., \cite{Li2019RLApp} and a blog.\footnote{\url{https://medium.com/@yuxili/rl-applications-73ef685c07eb}} 
(Both were last updated in 2019.)



\begin{figure}[h]
\includegraphics[width=1.0\linewidth]{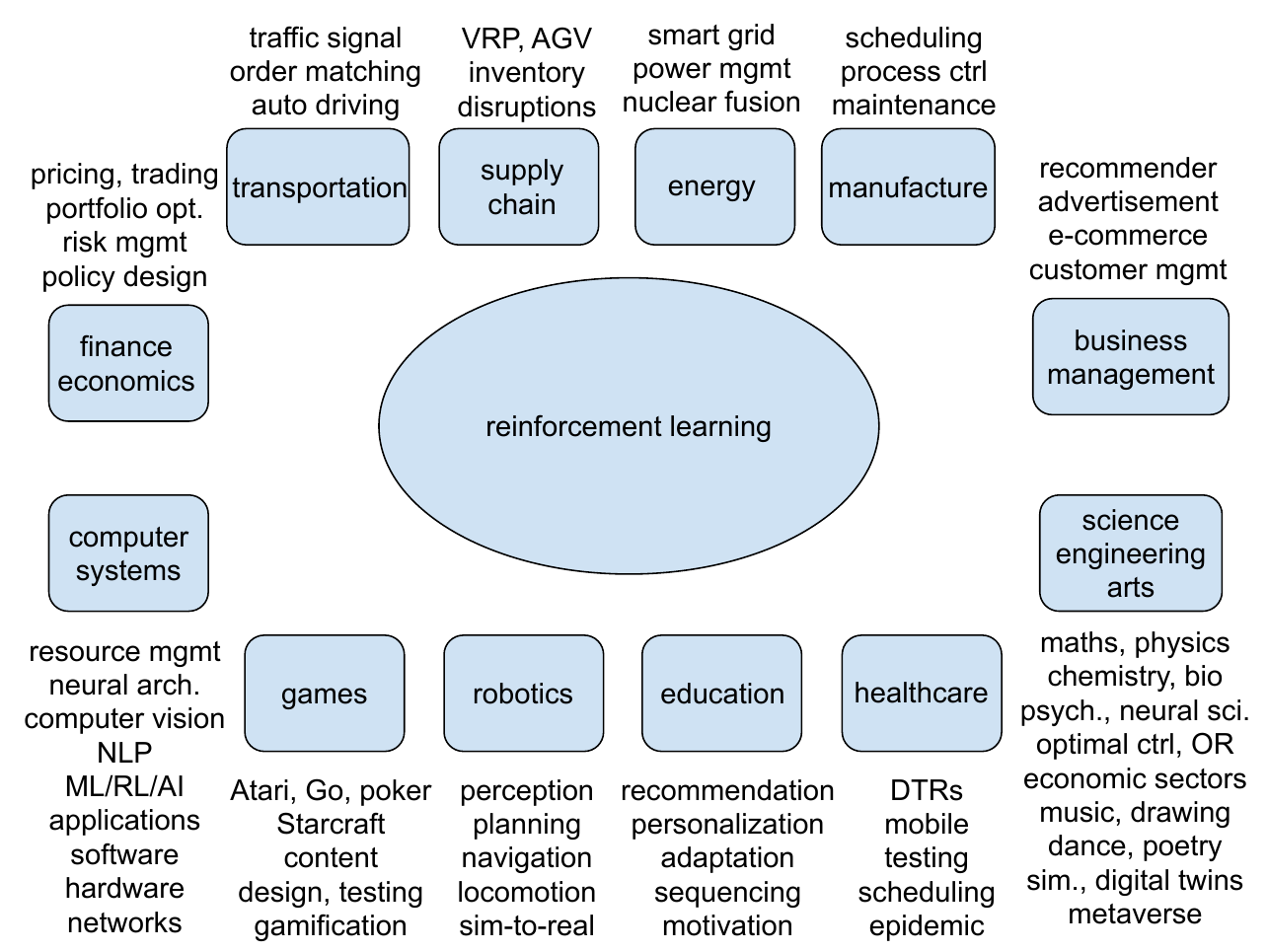}
\caption{Reinforcement Learning is Widely Applicable.}
\label{applications}
\end{figure}

\subsection{Production and Services}

Several RL applications have reached the stage of production and/or services.

The Microsoft Real World Reinforcement Learning Team created Decision Service~\citep{Agarwal2016}, and received the 2019 Inaugural ACM SIGAI Industry Award for Excellence in Artificial Intelligence.\footnote{\url{https://sigai.acm.org/awards/industry_award.html}. See the open source at \url{https://github.com/Microsoft/mwt-ds} and the webpage at \url{http://ds.microsoft.com}.}
Furthermore, Microsoft launched Personalizer as part of Azure Cognitive Services within Azure AI platform, with wide applications inside like in more Microsoft products and services, Windows, Edge browser, Xbox, and outside of Microsoft.\footnote{See a blog at \url{https://blogs.microsoft.com/ai/reinforcement-learning/}.}
Microsoft also launched Project Bonsai for building autonomous systems.\footnote{\url{https://www.microsoft.com/en-us/ai/autonomous-systems-project-bonsai}}

Google has applied RL in diverse application areas, like recommendation~\citep{Chen2019WSDM, Ie2019RL4RealLife}, chip design~\citep{Mirhoseini2021}, and AutoML, which attempts to make ML easily accessible. 
Google Cloud AutoML provides services like the automation of neural architecture search~\citep{Zoph2017, Zoph2018Transfer}, device placement optimization~\citep{Mirhoseini2017}, and data augmentation\citep{Cubuk2019AutoAugment}.
\cite{Co-Reyes2021} recently automate RL algorithms themselves.

Facebook/Meta  has open-sourced ReAgent~\citep{Gauci2019RL4RealLife}, an RL platform for products and services like notification delivery.\footnote{See the open source at \url{https://github.com/facebookresearch/ReAgent}.}
Didi has applied RL to ride sharing order dispatching~\citep{Qin2020Didi},
and won the INFORMS 2019 Wagner Prize Winner.
OpenAI recently proposes InstructGPT to fine-tune GPT-3 large language models with human feedback for user intent alignment~\citep{Ouyang2022}.
We discuss Meta's ReAgent and Didi's ride sharing in more details in later sections.

\subsection{Games}

We have witnessed breakthroughs in RL/AI recently, in particular, in games.

Atari games are for single-player games and single-agent decision making in general.
\citet{Mnih-DQN-2015} introduce Deep Q-Network (DQN), which ignited the current round of popularity of deep RL. 
Recently, Agent57~\citep{Badia2020} and Go-Explore~\citep{Ecoffet2021} achieve superhuman performance for all 57 Atari games.

Computer Go is a two-player perfect information zero-sum game, a very hard problem, pursued by many researchers for decades.
AlphaGo made a phenomenal achievement by defeating a world champion, and set a landmark in AI. 
\citet{Silver-AlphaGo-2016} introduce AlphaGo. 
\citet{Silver-AlphaGo-2017} introduce AlphaGo Zero, mastering the game of Go without human knowledge.
\citet{Silver-AlphaZero-2018} introduce AlphaZero, extending AlphaGo Zero to more games like chess and shogi (Japanese chess).
\cite{Tian2019OpenGo} reimplement AlphaZero, open source it, and report an analysis. 
The core techniques are deep learning, reinforcement learning,  Monte Carlo tree search (MCTS)\citep{Browne2012, Gelly12}, upper confidence bound for trees (UCT)\citep{Kocsis2006}, self play and policy iteration.
The series of AlphaGo, AlphaGo Zero, and AlphaZero have lifted the bars  so high for Go, chess, shogi, and likely many two-player perfect information zero-sum games that it is hard or impossible for human players to compete with AI programs.

Moreover, (deep) RL has shown its strength in multi-player and/or imperfect information games.
\cite{Moravcik2017} introduce DeepStack and \cite{Brown2017Science} introduce Libratus, for no-limit, heads-up Texas Hold'em Poker.
The core techniques are counterfactual regret minimization (CFR)~\citep{Zinkevich2007} and policy iteration.
\cite{Zhao2022AlphaHoldem} introduce AlphaHoldem and achieve high performance with end-to-end RL (without CFR).
DeepStack, Libratus and AlphaHoldem are for two-player imperfect information zero-sum games, a family of problems inherently difficult to solve. 
\cite{Brown2019Science} is an attempt for multi-player poker.
\cite{Schmid2021} propose to combine guided search, self play, and game-theoretic reasoning for both perfect and imperfect information games.

Deepmind AlphaStar defeats top human players at StarCraft II, a complex Real-Time Strategy (RTS) game~\citep{Vinyals2019AlphaStar}.
Deepmind has achieved human-level performance in Catch The Flag~\citep{Jaderberg2019}. 
OpenAI Five defeats good human players at Dota 2~\citep{OpenAI2019}. 
\cite{Baker2020HideSeek} show emergent strategies with a self-supervised autocurriculum in hide-and-seek. 
There are recent achievements in Hanabi~\citep{BARD2020103216, Lerer2020}, Diplomacy~\citep{Bakhtin2021}, DouDizhu~\citep{Jiang2019Dou, Zha2021Dou}, Mahjong~\citep{Li2020mahjong}, etc.
Such achievements in multi-player games show the progress in mastering tactical and strategical team plays. 



Games correspond to fundamental problems in AI, computer science, and many real life scenarios.
Each of the above represents a big family of problems and the underlying techniques can be applied to a large number of applications. 
Single agent decision making is by nature widely applicable.
AlphaGo series papers mention the following application areas:
general game-playing, classical planning, partially observed planning, scheduling, and constraint satisfaction.
DeepStack paper mentions:
defending strategic resources and robust decision making for medical treatment recommendations. 
Libratus paper mentions:
business strategy, negotiation, strategic pricing, finance, cybersecurity, military applications, auctions, and pricing.

It is straightforward to think about applying techniques in AlphaGo series to discrete problems. 
\cite{Bertsekas2022AlphaZero} discusses lessons from AlphaZero for continuous problems in optimal, model predictive, and adaptive control, and sheds light on the benefits of on-line decision making on top of off-line training.

The above is about game playing.
RL is \enquote{mature} for many single-, two- and multi-player games; or, game playing is the most mature application area of RL.
There are other aspects of games, e.g., game testing~\citep{Roohi2018, Zheng2019Wuji} and procedural content generation~\citep{Liu2021}. 
Sports can be regarded as games in real life.
\cite{Won2020} report a curling robot with human-level performance. 
\cite{Liu2021football} investigate simulated humanoid football from motor control to team cooperation.
\cite{Wurman2022racing} develop automobile racing agent, winning the world’s best e-sports drivers.

Games to AI is like fruit flies to genetics.
Games have the potential to make us better and change the world~\citep{McGonigal2011, Schell2020, Yannakakis2018}.

\subsection{Bandits}
\label{bandits}

Bandits appear as the most mature techniques in RL.\footnote{We put bandits as an opportunity due to its maturity in both theory and practice; while the other opportunities are about application areas.}
Contextual bandits are effective to select actions based on contextual information, 
e.g., in news article recommendation, 
selecting articles for users based on contextual information of users and articles, 
like historical activities of users and descriptive information and categories of content~\citep{Li2010}.

In the following, we discuss the Decision Service~\citep{Agarwal2016}, a software system based on contextual bandits developed in Microsoft, with general failures and principles in developing and deploying contextual bandits. 
We discuss software issues in Section~\ref{software}.


\subsubsection*{Decision Service}

The contextual bandits setting encounters two challenges.
1) Partial feedback. The reward is for the selected action only, but not for unexplored actions.
2) Delayed rewards. Reward information may arrive considerably later after taking an action.
As a result, an implementation usually faces the following failures: 
1) partial feedback and bias;
2) incorrect data collection; 
3) changes in environments; and
4) weak monitoring and debugging.

The machine learning methodology includes contextual decisions, exploration and logging, policy learning, non-stationarity and problem framing, partially addressing failures 1, 3 and 4.  Each interaction of contextual decisions follows a protocol: 1) a context arrives and is observed by the application, 2) the application chooses an action to take, and 3) a reward for the action is observed by the application. Exploration is used to randomize choices of actions, to balance the explore-exploit tradeoff, guaranteeing a minimum performance while exploring for better alternatives. Each exploration data point consists an observation, an action, a reward, and the exploration policy's probability of choosing the action given the observation.

A/B testing is an approach for exploration to address the issue of partial feedback.
It tests two policies by running them on random user traffic proportionally.
Thus the data requirements scale linearly with the number of policies.
Contextual bandits allows to test and optimize exponential number of policies with the same amount of data.
And it does not require policies to be deployed to test, to save business and engineering effort.
Supervised learning does not support exploration, so not tackling contextual settings or the issues of partial feedback and bias.

Non-stationarity is addressed with a continuous loop so that an online learning algorithm incorporates new data quickly, and the learning rate is periodically reset or kept constant to stay sensitive to new data. The problem framing defines the context/features, action set, and reward metric,
and is eased by feature auto-generation for content recommendation applications and the ability to test different problem framings without new data collection.

The Decision Service defines four system abstractions, namely, 
explore to collect data, log data correctly, learn a good model, and deploy the model in the application,
to complete the loop for effective contextual learning,
by providing the capacity of policy evaluation,
aiming to address all four failures.

\subsubsection*{More about Bandits}

Methods in bandits are foundational for exploration, which we will discuss in Section~\ref{exploration}.

\cite{Auer2002} introduce upper confidence bounds (UCB).
UCB is an approach to action selection according to an action's potential for being optimal, considering both the closeness of the estimates being maximal and the uncertainties or variance in those estimates.
\cite{Kocsis2006} introduce UCB applied to trees (UCT),
which is critical for the success of AlphaGo series via Monte Carlo tree search (MCTS).

\cite{Li2010} propose to solve personalized news articles recommendation with contextual bandits. 
\cite{Chapelle2011} present an empirical evaluation of Thompson Sampling.
\cite{Rome2021} present their experience with a contextual bandit model for troubleshooting notification in a chatbot.
\cite{Karampatziakis2019RL4RealLife} discuss lessons from real life RL in a customer support bot.

\cite{Lattimore2018} is a book about mathematical analysis of bandit algorithms, with guiding principles for algorithm design,  intuition for analysis, and empirical demonstrations.
\cite{Slivkins2019} is a book about multi-armed bandits.
\cite{Russo2018Thompson} presents a tutorial on Thompson sampling.
\cite{Glowacka2019} presents bandit algorithms in information retrieval.
\cite{Tewari2017} present a survey on contextual bandits in mobile health.
\cite{Bouneffouf2019} presents a survey on practical applications of bandits.

\subsection{Recommender Systems}

A recommender system can suggest products, services, information, etc. to users according to their preferences. It helps alleviate the issues with overwhelming information these days by personalized recommendations, e.g., for news, movies, music, restaurants, etc. 
Recommendation and personalization are powered by machine learning and data mining~\citep{Aggarwal2016}, by bandits~\citep{Lattimore2018, Li2010},
and more and more by general RL~\citep{Zhao2021Tutorial}.

\cite{Boutilier2019} discusses challenges for RL in recommender system.
In user-centric recommender systems, 
it is essential to understand and meet genuine user needs and preferences, 
using natural, unobtrusive, transparent interaction with the users.   
Recommender systems need to
estimate, ellicit, react to, and influence user latent state, e.g., satisfaction, transient vs. persistent preferences, needs, interests, patterns of activity etc.
by natural interaction with users,
and acting in user's best interests, i.e, user preferences, behavioural processes governing user decision making and reasoning, etc.
RL will play a central role in all of these.
User-facing RL encounters some challenges,
1) large scale with massive number of users and actions, multi-user (MDPs), and combinatorial action space, e.g., slates;
2) idiosyncratic nature of actions, like stochastic actions sets, dynamic slate sizes, etc.;
3) user latent state, leading to high degree of unobservability and stochasticity, w.r.t. preferences, activities, personality, user learning, exogenous factors, behavioral effects, etc, which add up to long-horizon, low signal noise ratio (SNR) POMDPs; 
and,
4) eco-system effects, e.g., incentives, ecosystem dynamics, fairness, etc.

\cite{Chen2019WSDM} propose to scale REINFORCE to a production top-K recommender system for YouTube with a large scale action space, with a top-K off-policy method to tackling the issues of biases due to learning from feedback logs generated from multiple behavior policies and recommendations with multiple items. 
\cite{Ie2019RL4RealLife} propose RecSim, a configurable recommender systems environment. 

\cite{Gauci2019RL4RealLife} present ReAgent,\footnote{It is renamed from Horizon. See \url{https://github.com/facebookresearch/ReAgent}.} 
Facebook’s open source applied RL platform, featuring data preprocessing, feature normalization, data understanding tool, deep RL model implementation, multi-node and multi-GPU training, counterfactual policy evaluation, optimized serving, and tested algorithms, considering real life issues like large datasets with varying feature types and distributions, high dimensional action spaces, slow feedback loop in contrast to simulation, and, more care for experiments in real systems. The authors also discuss cases of real life deployment of notifications at Facebook, including push notifications and page administrator notifications. 

\cite{Theocharous2020} discuss RL for strategic recommendations.
\cite{Dacrema2019} and \cite{Rendle2019} conduct critical analysis of recommender systems.
A major application area of contextual bandits is recommender systems, as we discussed in Section~\ref{bandits}, e.g.,\cite{Agarwal2016}, \cite{Li2010} and \cite{Karampatziakis2019RL4RealLife}.

In some sense, all RL problems are about recommendations: recommend an action/strategy for a given state/observation.
That is, recommendation is about decision making.
Here we take the narrow view of recommendations as discussed above, in particular, in Web and e-commerce applications.

\subsection{Robotics}
 
Robotics is a classical application area for RL (and optimal control)
and has wide applications, e.g., in manufacturing, supply chain, healthcare, etc.
Robotics pose challenges to RL,
including 
1) dimensionality, 
2) real-world examples, 
3) under-modeling (models not capturing all details of system dynamics) and model uncertainty,
and 4) reward and goal specification~\citep{Kober2013}. 
RL provides tractable approaches to robotics, 
1) through representation, including state-action discretization, value function approximation, and pre-structured policies;
2) through prior knowledge, including demonstration, task structuring, and directing exploration; 
and 3) through models, including mental rehearsal, which deals with simulation bias, real-world stochasticity, and optimization efficiency with simulation samples, and approaches for learned forward models~\citep{Kober2013}.

DL and RL have enabled rapid progress in robotics. 
 OpenAI trained Dactyl for a human-like robot hand to dextrously manipulate physical objects~\citep{OpenAI2018}.
A series of work has advanced the progress in quadrupedal robots locomotion over challenging terrains in the wild, integrating both exteroceptive and proprioceptive perception~\citep{Hwangbo2019, Lee2020quadrupedal, Miki2022}. 
\cite{Peng2018} present DeepMimic for simulated humanoid to perform highly dynamic and acrobatic skills.  
\cite{Peng2021AMP} propose to combine adversarial motion priors with RL for stylized physics-based character animation. 
\cite{Luo2021RSS} investigate policies for industrial assembly with RL and imitation learning, tackling the prohibitively large design space. 
\cite{Krishnan2021} introduce a simulator for resource-constrained autonomous aerial robots.
\cite{Wurman2022racing} develop automobile racing agent in simulation,  the PlayStation game Gran Turismo, to win the world’s best e-sports drivers.

\cite{Ibarz2021} review how to train robots with deep RL and discuss outstanding challenges and strategies to mitigate them:
1) reliable and stable learning;
2) sample efficiency: 2.1) off-policy algorithms, 2.2) model-based algorithms, 2.3) input remapping for high-dimensional observations, and 2.4) offline training; 
3) use of simulation: 3.1) better simulation, 3.2) domain randomization, and 3.3) domain adaptation;
4) side-stepping exploration challenges: 4.1) initialization, 4.2) data aggregation, 4.3) joint training, 4.4) demonstrations in model-based RL, 4.5) scripted policies, and 4.6) reward shaping; 
5) generalization: 5.1) data diversity and 5.2) proper evaluation;
6) avoiding model exploitation;
7) robot operation at scale: 7.1) experiment design, 7.2) facilitating continuous operation, and 7.3) non-stationarity owing to environment changes; 
8) asynchronous control: thinking and acting at the same time;
9) setting goals and specifying rewards;
10) multi-task learning and meta-learning;
11) safe learning: 11.1) designing safe action spaces, 11.2) smooth actions, 11.3) recognizing unsafe situations, 11.4) constraining learned policies, and 11.5) robustness to unseen observations; 
and 12) robot persistence: 12.1) self-persistence and 12.2) task persistence.

\cite{Abbeel2021} discusses that, similar to the pre-training then finetuning in computer vision on ImageNet and in NLP, like GPT-X and BERT, on Internet text, we might be able to pre-train large-scale neural networks for robotics as a general solution,  with unsupervised representation learning on Internet video and text, with unsupervised (reward-free) RL pre-training, mostly on simulators and little on the real world data, with human-in-the-loop RL, and with few shot imitation learning on demonstrations.

\cite{Mahmood2018} benchmark RL algorithms on real world robots.
\cite{Kroemer2021} review robot learning for manipulation.
\cite{Brunke2021Safe} survey safety issues in robotics.
\cite{Xiao2022motion} survey motion planning and control for robot navigation with machine learning.
Amazon has launched a physical RL testbed AWS DeepRacer, together with Intel RL Coach.
We discuss sim-to-real in Section~\ref{model}.

\subsection{Transportation}

Transportation is a critical infrastructure for everyday life and work, and it is related with supply chain and smart cities.
There are recent work in transportation like traffic light control and autonomous driving; see e.g. \cite{Haydari2022} for a survey. 
\cite{Wu2021Flow} present Flow, a benchmark for RL in traffic control.
\cite{Zhou2020SMARTS} present a simulation platform for multi-agent RL in autonomous driving.
\cite{Shang2021uplift} study partially observable environment estimation with uplift inference with an application in ride sharing.

Here we discuss applying RL to ridesharing~\citep{Qin2020Didi,Qin2021Didi}. 
Ridesharing order dispatching faces many challenges.
The supply and demand are dynamic and stochastic.
Short system response time and reliability are expected.
There are multiple business objectives.
For a driver-centric objective, we want to maximize the total income of the drivers on the platform;
whereas for a passenger-centric objective, we want to minimize the average pickup distance of all the assigned orders.
Marketplace efficiency metrics concern with response rate and fulfillment rate.
Production requirements and constraints consider
computational efficiency, 
system reliability, and
changing business requirements.
\cite{Qin2020Didi} discuss Didi's solution approaches to ridesharing order dispatching,
from the myopic yet practical combinatorial optimization,
to semi-Markov decision process,
tabular temporal difference learning,
deep (hierarchical) RL, and 
transfer learning. 
\cite{Qin2021Didi} discuss RL for operational problems in ridesharing, 
including: pricing, online matching, vehicle repositioning, route guidance (navigation), ride-pooling (carpool),  vehicle routing problem (VRP), model predictive control (MPC), data sets \& environments, 
and discuss challenges and opportunities, including
ride-pooling, joint optimization, heterogeneous fleet, simulation \& Sim2Real, non-stationarity, and business strategies.

The success of applying RL to transportations, in particular, to ridesharing, presents a good example for problems in operations research, like supply chain and smart grids, and in general, for problems related to networks, graphs, and combinatorial optimization, e.g., resource allocation in communication networks. 

\subsection{Finance and Economics}

RL is a natural solution to many sequential decision making problems in finance and economics, like option pricing, trading, and portfolio optimization, policy design, etc. 
Simulation based methods like RL are flexible to deal with complex, high dimensional scenarios. 

Options are fundamental financial instruments.
\cite{Longstaff01, Tsitsiklis01, Li2009Option} propose methods to calculate the conditional expected value of continuation, the key to American option pricing.
\citet{Hull14} provides an introduction to options and other financial derivatives.
\cite{Wang2019Stock} propose a buying winners and selling losers investment strategy with deep RL, leveraging the momentum phenomenon in finance.
\cite{Boukas2021} study continuous intraday market bidding.
\cite{Liu2020FinRL} introduce a deep RL library to develop stock trading strategies,
and \cite{Liu2021FinRL-Meta} build a universe of market environments for data-driven quantitative finance.\footnote{\url{https://github.com/AI4Finance-Foundation}}
\cite{Kearns2013} discuss machine learning for market microstructure and high frequency trading, in particular, order book execution.
\cite{Bacoyannis2018} discuss idiosyncrasies and challenges of data driven learning in electronic trading. 
\cite{Zheng2021AIEconomist} propose AI Economist to design optimal economic policies with deep RL to address the issues with counterfactual data, behavioral models, and evaluation of policies and behavioral responses.

There are two schools in finance: Efficient Markets Hypothesis (EMH) and behavioral finance~\citep{Lo04}. According to EMH, \enquote{prices fully reflect all available information} and are determined by the market equilibrium. However, psychologists and economists have found a number of behavioral biases that are native in human decision-making under uncertainty. 
For example, Amos Tversky and Daniel Kahneman demonstrate the phenomenon of loss aversion, in which people tend to strongly prefer avoiding losses to acquiring gains~\citep{Kahneman2011}. \citet{Prashanth2016} investigate prospect theory with RL. 
\citet{Lo04} proposes the Adaptive Market Hypothesis to reconcile EMH and behavioral finance, where the markets are in the evolutionary process of competition, mutation, reproduction and natural selection, where RL may play an important role. 

\subsection{Healthcare}


Personalized medicine systematically optimizes patients' health care, in particular, for chronic conditions and cancers using individual patient information, potentially from electronic health/medical record (EHR/EMR).  
Dynamic treatment regimes (DTRs) or adaptive treatment strategies are sequential decision making problems~\citep{Chakraborty2014}. 
\citet{LiuYao2018NIPS} study off-policy policy evaluation and its application to sepsis treatment.
\citet{KallusZhou2018NIPS} study confounding-robust policy improvement and its application to acute ischaemic stroke treatment.
\cite{Tomkins2021} learn personalized user policies from limited data and non-stationary responses to treatments, achieve a high probability regret bound, perform an empirical evaluation, and conduct a pilot study in a live clinical trial.
\citet{Gottesman2020} study interpretable RL by highlighting influential transitions and apply it to medical simulations and Intensive Care Unit (ICU) data.
\cite{Prasad2017} investigate an RL approach to weaning of mechanical ventilation in ICUs.
\cite{Menictas2019} discuss AI decision making in mobile healthcare with micro-randomized trials and just-in-time adaptive interventions (JITAIs).


RL is a promising framework to combat epidemics; and epidemics are a fruitful application area for RL to make substantial real life impact~\citep{
Li2020epidemics}.
In fact, RL has shown its utility in combating the ongoing pandemic COVID-19.
\cite{Bastani2021COVID19} propose to use bandits algorithms for COVID-19 tests in Greece.
\cite{Capobianco2021JAIR} study how to optimize mitigation policies considering both economic impact and hospital capacity.
\cite{Colas2021JAIR} propose a toolbox for optimizing control policies in epidemiological models.
\cite{Trott2021COVID19} propose to optimize economic and public policy, in particular, for COVID-19, with AI Economist~\citep{Zheng2021AIEconomist}.

\cite{Gottesman2019} present guidelines for RL in healthcare.
\cite{Wiens2019} discuss how to do no harm in the context of healthcare.
\cite{Tewari2017} present a survey on contextual bandits in mobile health.
\cite{Yu2023Health} present a survey about RL in healthcare.

\subsection{Education}

RL/AI can help education with recommendation, personalization, and adaptation, etc.
\cite{Cai2021edu} propose an educational conversational agent with rules integrated with contextual bandits for math concepts explanation, practice questions, and customized feedbacks. 
\cite{Doroudi2019} review RL for  instructional sequencing, and show that ideas and theories from cognitive psychology and learning sciences help improve performance.
\citet{Oudeyer2016} discuss theory and applications of intrinsic motivation, curiosity, and learning in educational technologies.

\cite{Singla2021} present opportunities and challenges for RL for education based on a recent workshop. 
The authors identify the following challenges: 
1) lack of simulation environments,
2) large or unbounded state space representations, 
3) partial observability of students' knowledge, 
4) delayed and noisy outcome measurements, 
and 5) robustness, interpretability, and fairness.
The authors list the following research directions:
1) personalizing curriculum across tasks,
2) providing hints, scaffolding, and quizzing,
3) adaptive experimentation and A/B testing,
4) human student modelling,
and 5) content generation.
See \url{https://rl4ed.org/edm2021/} for more details,
in particular, invited talks by Emma Brunskill and Shayan Doroudi among others.

\subsection{Combinatorial Optimization}
 
Combinatorial optimization is relevant to a large range of problems in operations research, AI and computer science, e.g., travelling salesman problem (TSP)~\citep{Vinyals2015}, vehicle routing problem (VRP)~\citep{Chen2019Combinatorial, Lu2020ICLR}, scheduling~\citep{Mao2019SIGCOMM}, network planning~\citep{Zhu2021SIGCOMM}, and ride-sharing~\citep{Qin2020Didi,Qin2021Didi} as discussed earlier. 
Many combinatorial optimization problems are NP-hard, and (traditional) algorithms follow three approaches: 
exact algorithms, approximate algorithms, and heuristics,
and all of which require specialized knowledge and human efforts for trail-and-error.
Machine learning can help in several ways~\citep{Bengio2021Combinatorial}.
In end-to-end learning, a solution is directly provided to the problem by applying the trained ML model to inputs instance, e.g., in~\cite{Vinyals2015}.
Machine learning can help to configure algorithms, e.g., for hyper-parameters like learning rate.
Machine learning can work alongside with optimization algorithms, e.g., selecting the branching variable in branch and bound methods, like \cite{Gasse2019}, and learning heuristics to improve the solution, like~\cite{Chen2019Combinatorial} and~\cite{Lu2020ICLR}.

\subsection{Computer Systems} 

RL has been applied to the whole spectrum of computer systems, from the very bottom hardware, to system softwares, to RL/ML/AI themselves, to networking systems, to various applications, e.g.,
chip design~\citep{Mirhoseini2021}, 
neural architecture search~\citep{Zoph2017},
compiler~\citep{Cummins2021},
cluster scheduling~\citep{Mao2019SIGCOMM}, 
network planning~\citep{Zhu2021SIGCOMM},
device placement~\citep{Mirhoseini2017},
data augmentation~\citep{Cubuk2019AutoAugment},
database management system (DBMS)~\citep{Zhang2019SIGMOD},
software testing~\citep{Zheng2019Wuji},
conversational AI~\citep{Gao2019Neural},
natural language processing (NLP)~\citep{Wang2018ACL},
computer vision~\citep{Lu2019CVPR},
and automating RL algorithms themselves~\citep{Co-Reyes2021, Guez2018, Kirsch2020}.
\cite{Mao2019NeurIPS} present Park, an open platform for learning augmented computer systems, covering a wide range of systems problems.
See \cite{Luong2019networking} for a survey about applications of deep RL in communications and networking.
Computers are ubiquitous, so in some sense computer systems can cover all topics we discuss in this article. 

 
\subsection{Science and Engineering}

A problem in natural science and engineering may come with a clear objective function straightforward to evaluate, in particular, when comparing with problems in social sciences. 
The scientific and engineering understandings can help build effective inductive priors, improve search efficiency, and/or construct models/simulators. 
Here we list some examples.
\cite{Todorov2012} propose the physics engine Mujoco for model-based control.
\cite{Seth2018OpenSim} propose OpenSim for human and animal movement with musculoskeletal dynamics and neuromuscular control. 
Habitat 2.0~\citep{Szot2021Habitat}, 
Isaac Gym~\citep{Makoviychuk2021Isaac}, and 
ThreeDWorld~\citep{Gan2021ThreeDWorld}
provide interactive multi-modal physical simulation platforms.

Many problems in science and engineering, including topics we discuss earlier, like robotics and transportation, as well as healthcare, finance, and economics,  are traditionally modelled as optimal control  or operations research problems, or generally, dynamic programming or (stochastic) optimization problems.
See Section~\ref{foundation} for more discussion.

\cite{Degrave2022Nature} report an extraordinary attainment applying deep RL to nuclear fusion, promising for sustainable energy.
RL has been applied to chemical retrosynthesis~\citep{Segler2018} and drug discovery~\citep{Popova2018, Zhavoronkov2019}.
\cite{Bengio2021GFlowNet} introduce Generative Flow Networks (GFlowNets) to sample approximately in proportion to a given reward function, with applications in drug discovery.
\citet{Lazic2018NIPS} from Google AI study data center cooling, a real-world physical system, with model-predictive control (MPC) and RL.
\cite{Zhan2021DeepThermal}  propose to optimize combustion for thermal power generating units with offline RL.
\cite{Henry2021ANM} propose an RL environment Gym-ANM for active network management tasks in electricity distribution systems.

\cite{Anderson2011} discuss smart grid.
\cite{Chen2021power} review RL in power systems.
\cite{Shin2019process} and \cite{Nian2020process} review RL in process control.
\cite{Kiumarsi2018} review RL in optimal and autonomous control.

\section{Challenges}

\begin{figure}[h]
\centering
\includegraphics[width=0.96\linewidth]{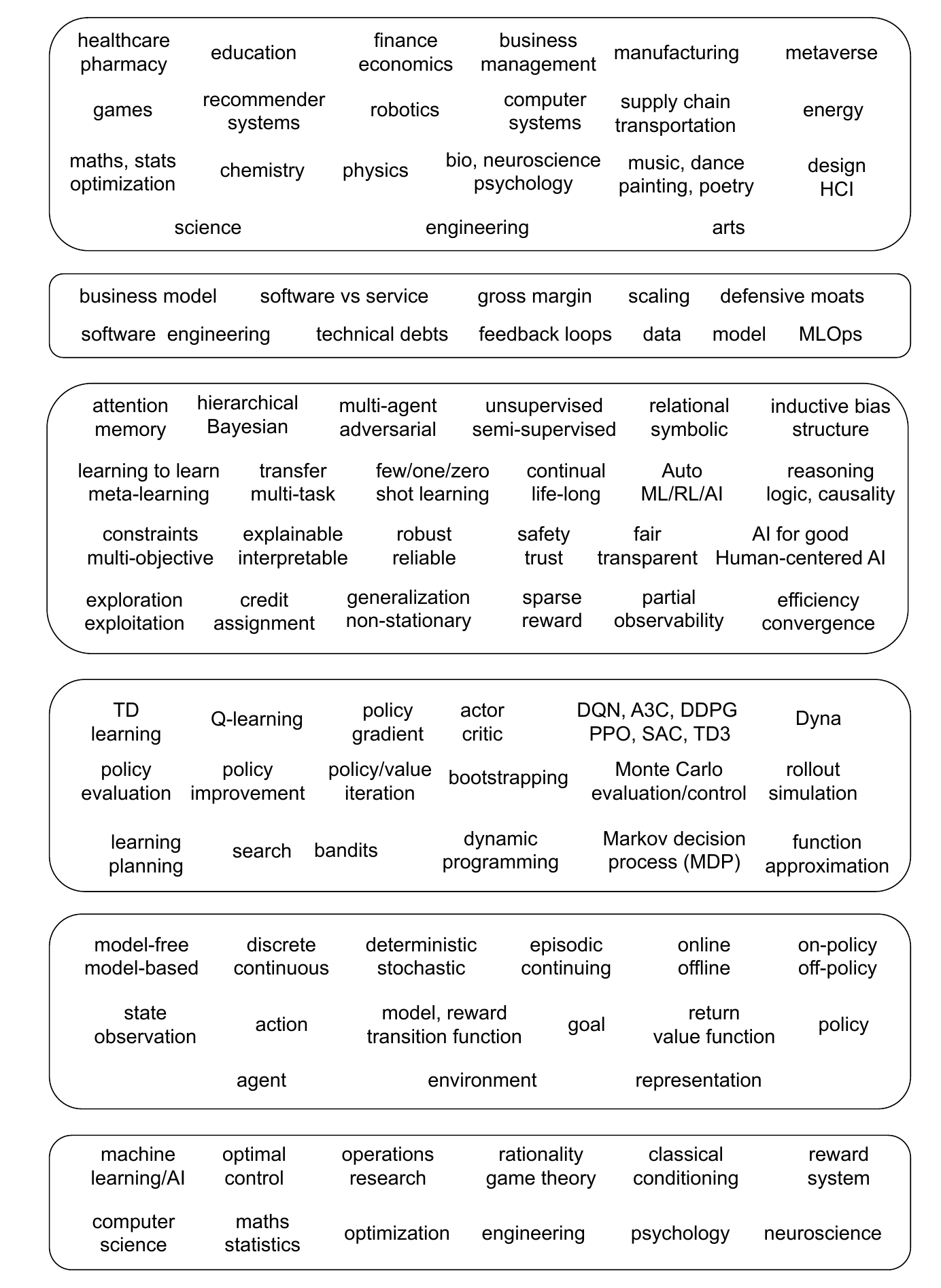}
\caption{Challenges of reinforcement learning. From the bottom to the top, roughly: the foundational disciplines for RL, essential RL elements, core solution methods for RL problems with typical algorithms, theoretical and algorithmic issues and advanced mechanisms for RL solutions, software and business issues, and various applications for RL. (The part for software and business calls for more research and development.)}
\label{issues}
\end{figure}

Although RL has made significant progress, there are still many issues. 
Sample efficiency, credit assignment, exploration vs. exploitation, and representation are common issues.
Value function approaches with function approximation, in particular with DNNs, encounters the deadly triad, i.e., instability and/or divergence caused by the integration of off-policy, function approximation, and bootstrapping.
Reproducibility is an issue for deep RL, i.e., experimental results are influenced by hyperparameters, including network architecture and reward scale, random seeds and trials, environments, and codebases~\citep{Henderson2018}.
Reward specification may cause problems, and a reward function may not represent the intention of the designer.
Researchers and practitioners are still identifying issues, e.g., delusional bias~\citep{LuTyler2018NIPS} and expressivity of Markov reward~\citep{Abel2021reward}, and address them.
Figure~\ref{issues} illustrates challenges of reinforcement learning.



\cite{Dulac-Arnold2021} identify nine challenges for RL to be deployed in real-life scenarios, namely,
learning on the real system from limited samples,
system delays,
high-dimensional state and action spaces,
satisfying environmental constraints,
partial observability and non-stationarity,
multi-objective or poorly specified reward functions,
real-time inference,
offline RL training from fixed logs, and
explainable and interpretable policies.
Practitioners report lessons in RL deployment, e.g., 
\cite{Rome2021}, 
\cite{Karampatziakis2019RL4RealLife}.

Bearing in mind that there are many issues, there are efforts to address all of them, and RL is an effective technique for many applications.
As discussed by Dimitri Bertsekas, 
on one hand, there are no methods that are guaranteed to work for all or even most problems; 
on the other hand, 
there are enough methods to try with a reasonable chance of success for most types of optimization problems: deterministic, stochastic, or dynamic ones, discrete or continuous ones, games, etc.~\citep{Bertsekas2019slides}

\subsection{Foundation}
\label{foundation}

It is critical to have a sound foundation. 
Theory can usually guide practice. 
Once the foundation is right, RL will be everywhere.

\subsubsection*{What is Reinforcement Learning?}

It is essential to discuss what is RL, before discussing what is the foundation for RL.
RL is rooted in optimal control (in particular, dynamic programming and Markov decision process), learning by trail and error, and temporal difference learning, the latter two being related to animal learning. 
RL has foundation in theoretical computer science/machine learning, classical AI, optimization, statistics, optimal control, operations research, neuroscience, and psychology. 
(At the same time, RL also influences these disciplines.)    
RL has both art and science components. Rather than making it pure science, we need to accept that part of it is art.

The following are literal quotes from Section 1.6 in \cite{Sutton2018}.
\enquote{Reinforcement learning is a computational approach to understanding and automating goal-directed learning and decision making. It is distinguished from other computational approaches by its emphasis on learning by an agent from direct interaction with its environment, without requiring exemplary supervision or complete models of the environment. In our opinion, reinforcement learning is the first field to seriously address the computational issues that arise when learning from interaction with an environment in order to achieve long-term goals.} 
\enquote{Reinforcement learning uses the formal framework of Markov decision processes to define the interaction between a learning agent and its environment in terms of states, actions, and rewards. This framework is intended to be a simple way of representing essential features of the artificial intelligence problem. These features include a sense of cause and effect, a sense of uncertainty and nondeterminism, and the existence of explicit goals.}


\cite{Sutton2022model} proposes the common model of the intelligent agent for decision making across psychology, artificial intelligence, economics, control theory, and neuroscience, with the terminology of agent, world, observation, action and reward.

Richard Sutton discusses briefly about 
David Marr's three levels at which any information processing machine must be understood: computational theory, representation and algorithm, and hardware implementation.\footnote{\url{https://www.youtube.com/watch?v=hcJNFdZit-Q}}
AI has surprisingly little computational theory. 
The big ideas are mostly at the middle level of representation and algorithm.
RL is the first computational theory of intelligence.
RL is explicitly about the goal, the whats and whys of intelligence.

RL is a set of sequential decision problems.
It is not a specific set of algorithmic solutions, like TD, DQN, PPO, evolutionary search, random
search, semi-supervised learning, etc.
RL has three basic problems and many variations: 
online RL,
offline / batch RL, and 
planning/simulation optimization. 
See \cite{Szepesvari2020myths}.

RL methods deal with sampled, evaluative and sequential feedback, three weak forms of feedback  simultaneously, compared with exhaustive, supervised and one-shot feedback~\citep{Littman2015}.   

\subsubsection*{How to Achieve Stronger Intelligence?}

\cite{Sutton2019Bitter} argues that \enquote{The biggest lesson that can be read from 70 years of AI research is that general methods that leverage computation are ultimately the most effective, and by a large margin.}
This is supported by recent achievements in chess, computer Go, speech recognition, and computer vision.
We thus should leverage the great power of general purpose methods: search and learning.
Since the actual contents of minds are very complex,
we should build in only the meta-methods that can find and capture the arbitrary complexity,
and let our methods search for good approximations, but not by us.

\cite{Brooks2019Better} argues that we have to take into account the total cost of any solution, and that so far they have all required substantial amounts of human ingenuity.
As an example, Convolutional Neural Networks (CNNs) were designed by humans to manage translational invariance.
However, CNNs still suffer from issues like color constancy, by mistakenly recognizing a traffic stop sign with some pieces of tape on it as a 45 mph speed limit sign.
Network architecture design and special purpose computer architecture need human analysis.
The current AI still need massive data sets, huge amount of computation, and huge power consumption.
Furthermore, Moore’s Law slows down and Dennard scaling breaks down.   

\cite{Kaelbling2019Engineer} argues that it is constructive to specify the objective, which may be building a practical system in short or long term, understanding intelligence at the level of neural computation or behavior and cognition, or developing mathematical and computational theories.
The author's goal is to design software for intelligent robots.
It is somewhat a software engineering problem: given a spec, design the software.
However, the more abstract the spec, the more difficult the engineering problem.
To craft special-purpose programs with a general-purpose engineering methodology,
machine learning is critical to build robots in the factory and to learn to adapt in the wild.
When data and computation are not free, relatively general biases like structures, algorithms and reflexes help achieve sample efficient learning.

\cite{Szepesvari2020myths} argues that more data and less structure wins is a fallacy.
It is from the history of AI, which, however, is still too short.
The actual bitter lesson is: We can’t predict what works, what won't work. 
We won't run exhaustive search.
A resolution is to diversify: more data, less structure, smarter algorithms, etc.
Structure is always present, e.g., in gradient descent and neural architecture.
It is important to study structure.

\cite{Kaelbling2020} argues that the foundation for (robot) learning entails 
sample efficiency, generalization, composition,  and incremental learning,
which can be achieved by providing proper inductive biases, 
and meta-learning is an approach.

\cite{Botvinick2019Review} argue that sample inefficiency comes from two sources: 
incremental parameter adjustment to maximize generalization and avoid overwriting the effects of earlier learning, and weak inductive bias. 
The authors propose to address such slowness with episodic memory, meta-learning and their integration, and make the connection to fast and slow learning,
which is an idea in psychology underlying a Nobel prize in economics~\citep{Kahneman2011, Bengio2020}. 

\citet{Lake2017} discuss that the following are key ingredients to achieve human-like learning and thinking: 
1) developmental start-up software, or cognitive capabilities in early development, including:
1.1) intuitive physics and   
1.2) intuitive psychology;
2) learning as rapid model building, including:
2.1) compositionality,
2.2) causality, and
2.3) learning to learn;
3) thinking fast, including:
3.1) approximate inference in structured models and
3.2) model-based and model-free RL.
\cite{Brynjolfsson2022} discusses promise \& peril of human-like AI.

\subsubsection*{RL vs. Optimal Control and Operations Research}

Traditionally optimal control (OC)  and operations research (OR)  require a perfect model for the problem formulation.
RL can work in a model-free mode, and can handle very complex problems via simulation, which may be intractable for model-based OC or OR approaches.
AlphaGo, together with many games, are huge problems.
RL with a perfect model can return a perfect solution, in the limit.
RL with a decent simulator or with decently sufficient amount of data can provide a decent solution.  
Admittedly, there are adaptive and/or approximate approaches in OC and OR working with imperfect models, and/or for large-scale problems.
We see that RL, optimal control, and operations research are merging and advancing together, which is beneficial and fruitful for all these scientific communities.

\cite{Sutton2018} discuss the relationship between RL and optimal control in Section 1.7,
where we quote the following verbatim. 
\enquote{We consider all of the work in optimal control also to be, in a sense, work in reinforcement learning. We define a reinforcement learning method as any effective way of solving reinforcement learning problems, and it is now clear that these problems are closely related to optimal control problems, particularly stochastic optimal control problems such as those formulated as MDPs. Accordingly, we must consider the solution methods of optimal control, such as dynamic programming, also to be reinforcement learning methods. Because almost all of the conventional methods require complete knowledge of the system to be controlled, it feels a little unnatural to say that they are part of reinforcement learning. On the other hand, many dynamic programming algorithms are incremental and iterative. Like learning methods, they gradually reach the correct answer through successive approximations. As we show in the rest of this book, these similarities are far more than superficial. The theories and solution methods for the cases of complete and incomplete knowledge are so closely related that we feel they must be considered together as part of the same subject matter.}

\cite{Bertsekas2019, Bertsekas2019slides} presents ten key ideas for RL and optimal control:
1) principle of optimality,
2) approximation in value space,
3) approximation in policy space,
4) model-free methods and simulation,
5) policy improvement, rollout, and self-learning
6) approximate policy improvement, adaptive simulation, and Q-learning,
7) features, approximation architectures, and deep neural nets,
8) incremental and stochastic gradient optimization,
9) direct policy optimization: a more general approach, and
10) gradient and random search methods for direct policy optimization.

\cite{Powell2019} regards RL as sequential decision problems,
and identifies four types of policies: 
1) policy function approximations, e.g., parameterized policies,
2) cost function approximations, e.g., upper confidence bounding (UCB),
3) value function approximations, e.g., Q-learning, and
4) direct lookahead, e.g., Monte Carlo tree search (MCTS),
claiming they are universal, i.e., 
\enquote{any policy proposed for any sequential decision problem will consist of one of these four classes, and possibly a hybrid}.
\cite{Powell2019} proposes a unified framework based on stochastic control. 
 

\cite{Szepesvari2020myths} discusses that RL is not the answer to everything and RL is not the set of popular methods like TD, DQN and PPO. It is normal for certain methods like model predictive control (MPC) and random search to perform well on certain tasks. Methods have strengths and weaknesses. For a certain task, besides RL, we should consider other methods. 

\cite{Bertsekas2022AlphaZero} discusses lessons from AlphaZero for continuous problems in optimal, model predictive, and adaptive control.
\cite{Levine2018inference} regards RL and control as probabilistic inference.
\cite{Recht2019control} surveys RL from the perspective of optimization and control.

There are several recent books at the intersection of RL and optimal control and/or operations research, e.g.~\cite{Bertsekas2019, Powell2021, Meyn2022}. 

\subsubsection*{More Food for Thoughts}

We list some relevant materials below:
\begin{itemize}
\item \cite{Szepesvari2020myths} presents myths and misconceptions in RL.
\item \cite{Agarwal2020} present the FOCS 2020 Tutorial on the Theoretical Foundations of Reinforcement Learning. 
\item \cite{Krishnamurthy2021} present the COLT 2021 Tutorial on Statistical Foundations of Reinforcement Learning.
\item \cite{Schuurmans2019}, \cite{LiLihong2019} and \cite{Nachum2020duality} discuss optimization in RL.
\item \cite{Foster2021Complexity} analyze statistical complexity of interactive decision making.
\item Theory of Reinforcement Learning Boot Camp at Simons Institute at \url{https://simons.berkeley.edu/workshops/rl-2020-bc}.
\item \cite{Mitchell2021} discusses why AI is harder than we think.
\item \cite{Fortnow2022} discusses fifty years of P vs. NP and the possibility of the impossible with the progress in AI.
\end{itemize}

\subsection{Representation}



Here we discuss \enquote{representation} in a broad perspective, 
\footnote{Traditionally, the \enquote{representation} in \enquote{representation learning} basically refers to the \enquote{feature} in \enquote{feature engineering}~\citep{Bengio2013}.
On \url{https://iclr.cc}, it says \enquote{... the advancement of the branch of artificial intelligence called representation learning, but generally referred to as deep learning.}
}
i.e., it is relevant not only to the \enquote{feature} as in function approximation for state/observation, action, value function, reward function, transition probability, but also to agent, environment, 
problem representation, like MDP, POMDP, and contextual decision process (CDP)~\citep{Jiang2017},
and for representing knowledge, reasoning, and human intelligence, 
or basically any element in an RL problem.
We attempt to use such a notion of \enquote{representation} to unify the roles of (deep) learning in various aspects of (deep) RL.

When the problem is small, a tabular representation suffices to accommodate both states and actions.
For large-scale problems, we need function approximation to avoid the curse of dimensionality.
One approach is linear function approximation, using basis functions such as polynomial bases, tile-coding, radial basis functions, and Fourier basis~\citep{Sutton2018}.
Recently, nonlinear function approximations, in particular, DNNs, show exciting achievements. 
Common neural network structures and models include 
MLP, CNNs, RNNs, in particular LSTM and GRU, transformers, generative adversarial networks (GANs), and (variational) auto-encoder, etc.
There are new neural network architectures customized for RL problems, e.g., value iteration networks (VIN)~\citep{Tamar2016}, value prediction network (VPN)~\citep{Oh2017VPN}, and MCTSnets~\citep{Guez2018}.

General Value Functions (GVFs) learn, represent, and use knowledge of the world~\citep{Sutton2011}.
The aim of Universal Value Function Approximators (UVFAs) is to exploit structure across both states and goals, by taking advantages of similarities encoded in the goal representation, and the structure in the induced value function~\citep{Schaul2015}.  
\cite{Andrychowicz2017} propose Hindsight Experience Replay (HER) to combat with the sparse reward issue, i.e., after experiencing some episodes, to store every transition in the replay buffer, with not only the original goal for this episode but also some other goals. 
A value distribution is the distribution of the random return received by an RL agent~\citep{Bellemare2017Distributional, Bellemare2022Distributional}.
Successor representation is an approach for state distributions, which is related to value function.

Hierarchical representations, such as options~\citep{Sutton1999}, handle temporal abstraction.
Multi-agent RL model interactions among agents~\citep{Hernandez-Leal2019, Zhang2021MARL}.
Relational RL~\citep{Zambaldi2019} integrates statistical relational learning and reasoning with RL to handle entities and relations.


Unsupervised learning takes advantage of the massive amount of data without labels.
Self-supervised learning is a special type of unsupervised learning,
in which labels are created from data, although not given.
Unsupervised auxiliary learning~\citep{Jaderberg2017, Mirowski2017}, 
Horde~\citep{Sutton2011}, 
and contrastive unsupervised representations for RL~\citep{Srinivas2020}
are example approaches taking advantages of non-reward training signals in environments. 
\cite{Srinivas2021} present a tutorial on self(un)-supervised representation learning to improve RL, in particular, with contrastive learning~\citep{Chen2020SimCLR, Henaff2020CPC}.

There are renewed interests in deploying or designing networks for reasoning.  
\cite{Battaglia2018GN} propose \enquote{graph networks} as a building block for relational inductive biases, to learning entities, relations, and how to compose them for relational reasoning and combinatorial generalization.
\cite{Lamb2020GNN} discuss the integration of graph neural networks (GNNs) and neural-symbolic AI.
See \cite{Wu2021GNN} for a survey on GNNs.
There are discussions about addressing issues of current machine learning with causality~\citep{Pearl2018why}, and incorporating more human intelligence into artificial intelligence~\citep{Lake2017}.
\cite{Scholkopf2021} propose causal representation learning for out-of-distribution generalization. 

Although there have been enormous efforts for representation, 
since RL is fundamentally different from supervised learning and unsupervised learning, 
an optimal representation for RL is likely different from generic CNNs, RNNs and transformers,
thus it is desirable to search for an optimal representation for RL.
A hypothesis is that this would follow a holistic approach, by considering perception and control together,
e.g., in interactive perception~\citep{Bohg2017},
rather than treating them separately, e.g., by deciding a CNN to handle visual inputs, 
then fixing the network and designing a learning procedure to find optimal weights for value function and/or policy.
For example, \cite{Srinivas2018} propose plannable representation for goal-directed tasks with gradient-based planning.
\cite{Srinivas2021} discuss unsupervised learning for RL.
\cite{Abbeel2021} discusses a general solution, in particular, pre-training, for robotics.
However, note that \cite{Schuurmans2020} discusses reductionism in RL.



\subsection{Reward}

Rewards provide evaluative feedbacks for RL agents to make decisions.
\cite{Silver2021} present the  Reward-is-Enough hypothesis.

Rewards may be so sparse that it is challenging for learning algorithms, 
e.g., in computer Go, a reward occurs at the end of a game.  
Hindsight Experience Replay (HER)~\citep{Andrychowicz2017} is a way to handle sparse rewards.
Unsupervised auxiliary learning~\citep{Jaderberg2017} is an unsupervised way harnessing environmental signals.  
Intrinsic motivation~\citep{Barto2013, Singh2010} is a way to provide intrinsic rewards.
\cite{Colas2020} present a short survey for intrinsically motivated goal-conditioned RL.
\cite{Srinivas2021} present a tutorial on reward-free RL.

Reward shaping is to modify reward function to facilitate learning while maintaining optimal policy~\citep{Ng1999}. 
It is usually a manual endeavour.
\citet{Jaderberg2018Quake} employ a learning approach in an end-to-end training pipeline.  

Reward functions may not be available for some RL problems.
In imitation learning~\citep{Osa2018},  an agent learns to perform a task from expert demonstrations, with sample trajectories from the expert, without reinforcement signals.
Two main approaches for imitation learning are behavioral cloning and inverse RL.
Behavioral cloning, or learning from demonstration, maps state-action pairs from expert trajectories to a policy, maybe as supervised learning, without learning the reward function~\citep{Levine2021course}. 
Inverse RL is the problem of determining a reward function given observations of optimal behavior~\citep{Ng2000}. 
Probabilistic approaches are developed for inverse RL with maximum entropy~\citep{Ziebart2008} to deal with uncertainty in noisy and imperfect demonstrations.
\citet{Ross2010} reduce imitation learning and structured prediction to no-regret online learning, and propose Dataset Aggregation (DAGGER), which requires interaction with the expert. 
\cite{Belogolovsky2021} study a contextual inverse RL problem.
\cite{Likmeta2021} study inverse RL with multiple experts and non-stationarity.
 
A reward function may not represent the intention of the designer.
A negative side effect of a misspecified reward refers to potential poor behaviors resulting from missing important aspects.
An old example is about the wish of King Midas, that everything he touched, turned into gold. Unfortunately, his intention did not include food, family members, and many more.
\citet{Hadfield-Menell2016} propose a cooperative inverse RL (CIRL) game for the value alignment problem.
\cite{Dragan2020} talks about optimizing intended reward functions. 
  
\cite{Christiano2017} propose the approach of RL from human feedback, i.e., by defining a reward function with preferences between pairs of trajectory segments, to tackle the problems without well-defined goals and without experts' demonstrations, and to help improve the alignment between humans' values and the objectives of RL systems.
\cite{Lee2021PEBBLE} present unsupervised pre-training and preference-based learning via relabeling experience, to improve the efficiency of human-in-the-loop feedbacks with binary labels, i.e. preferences, provided by a supervisor.
\cite{Ouyang2022} propose to fine-tune large language models GPT-3 with human feedback, in particular, with RL~\citep{Christiano2017}, to follow instructions for better alignment with humans' values.  
 \cite{Wirth2017preference} present a survey of preference-based RL methods.
\cite{Zhang2021human} survey human guidance for sequential decision‐making.

From self-motivation theory~\citep{Ryan2020education}, the basic psychological needs of autonomy, competence and relatedness mediate positive user experience outcomes such as engagement, motivation and thriving~\citep{Peters2018}.
Flow is about the psychology of optimal experience~\citep{Csikszentmihalyi2008}. 
As such, they constitute specific measurable parameters for which designers can design in order to foster these outcomes within different spheres of experience.
Such self-motivation theory and flow, or positive psychology, may help the design of reward and human-computer interaction (HCI), and there are applications in games~\citep{Tyack2020}, education~\citep{Ryan2020education}, etc.
\cite{Cruz2020design} survey design principles for interactive RL.
It appears that self-motivation theory is under-explored in RL/AI.

\subsection{Exploration}
\label{exploration}

Exploration vs exploitation is a fundamental tradeoff for RL algorithms. 
Decision making under uncertainty faces a central issue: 
either exploit the information collected so far to make the currently best decision or explore for more information.

An RL agent usually uses  exploration to reduce its uncertainty about the reward function and transition probabilities of the environment. In tabular cases, this uncertainty can be quantified as confidence intervals or posterior of environment parameters, which are related to the state-action visit counts. 
In large-scale problems, this is approximated, e.g.,  \citet{Bellemare2016} and \citet{Tang2017}.
Intrinsic motivation~\citep{Barto2013, Schmidhuber2010, Oudeyer2016} suggests exploring based on the concepts of curiosity and surprise, so that actions will transition to surprising states that may cause large updates to the environment dynamics models. 
One particular example of measuring surprise is by the change in prediction error in learning process~\citep{Schmidhuber1991}, as studied recently in \citet{Bellemare2016}. 

There are several principles in trading off between exploration and exploitation, namely,
naive methods, 
optimism in the face of uncertainty, including,
optimistic initialization,
upper confidence bounds, 
probability matching, 
and information state search~\citep{Silver2015Course}.
These are developed in the settings of multi-armed bandits, 
but are applicable to RL problems.
We discuss bandits in Section~\ref{bandits}.
 
\citet{Levine2021course} discusses three classes of exploration methods in deep RL:
(1) optimistic exploration methods, 
which estimate state visitation frequencies or novelty,
typically with exploration bonuses, e.g.,  upper confidence bound (UCB)~\citep{Auer2002}, pseudo-count~\citep{Bellemare2016} and counting with hash-table~\citep{Tang2017}; 
(2)  Thompson sampling methods,
which learn distribution over Q-functions or policies,
then act according to samples, e.g., \citet{Osband2016};
and (3) information gain methods,
which reason about information gain from visiting new states, e.g., \citet{Houthooft2016}.
All these three methods follow the principle of optimism in the face of uncertainty.
 
\cite{Amin2021explore}  categorize exploration algorithms along two dimensions: 
reward-free vs. reward-based (whether extrinsic rewards are used in the action selection) and memory-free vs. memory-based (whether agent’s memory of the observed space are used in the action selection).
Blind exploration methods are reward-free and memory-free.
Intrinsic motivation methods are reward-free and memory-based.
Randomized action selection methods are reward-based and memory-free, and include value-based or policy-search based exploration (selecting actions based on the estimated value functions/rewards or the learned policies).
Reward-based and memory-based methods include deliberate exploration, probability matching, and optimism/bonus based.  
Optimism/bonus-based exploration follows the principle of optimism in the face of uncertainty, preferring actions with uncertain values, by adding a bonus to the extrinsic reward. It includes optimistic initialization, count-based, and prediction-error based methods.
Deliberate exploration operates by optimizing the exploration-exploitation tradeoff, and includes Bayes-adaptive and meta-learning based methods.
Probability matching, also known as Thompson Sampling~\citep{Russo2018Thompson}, selects an action according to the probability of being optimal.

\cite{Yang2021explore} categorize exploration algorithms along two dimensions: 
single-agent vs. multi-agent and uncertainty-oriented vs. intrinsic-motivation-oriented, 
and identify the following challenges: 
1) large state-action space,
2) sparse, delayed rewards,
3) white-noise problem, and
4) multi-agent exploration, due to exponential increase of the state-action space, coordinated exploration, and local- and global-exploration balance.

\cite{Li2012exploration} surveys theoretical approaches to exploration efficiency in RL.
\cite{Weng2020exploration} is a recent blog.
 
 

\subsection{Model, Simulation, Planning, and Benchmarks}
\label{model}

\cite{LeCun2022Autonomous} discusses autonomous AI, including world models.

An RL model refers to the transition probability and the reward function, mapping states and actions to distributions over next states and expected rewards, respectively. 
A model or a simulator specifies how an agent interacts with an environment.
A dataset may be collected offline, or generated by the dynamics rules, a model, or a simulator.
A model may be built from a dataset by estimating parameters, and/or with prior (physics) knowledge, or by a generative approach like GANs. 
A simulator may be built based on a model, explicitly, e.g., from game rules like the Arcade Learning Environment for Atari games~\citep{Bellemare2013, Machado2018ALE} or computer Go, chess and Shogi~\citep{Silver-AlphaZero-2018}, and physics like Mujoco~\citep{Todorov2012}, or implicitly, e.g. those with generative models~\citep{Ho2016, Chen2019ICML, Shi2019Taobao, Gui2020GAN}.
A benchmark includes datasets and/or simulators, together with algorithm implementations and evaluation methods.
Performance evaluation is essential for learning, engineering, and science; see e.g., \cite{Agarwal2021}. 
Planning works with a model, and may take a form of single agent search like A*, alpha-beta tree search, MCTS, or dynamic programming like value iteration.

Dyna is an early work that learns from both online and simulated data.
Value prediction network (VPN)~\citep{Oh2017VPN},
SimPLe~\citep{Kaiser2020SimPLe}, and
Dreamer~\citep{Hafner2020Dreamer}
are some recent work in model-based RL.
\cite{Schrittwieser2020MuZero} study how to do planning in Atari games, Go, chess, shoji with a learned model following AlphaZero.
\cite{Szepesvari2020misspecification} discusses model misspecification of RL.
\cite{Mordatch2020model} present the ICML 2020 tutorial on model-based methods in RL.

It is easier to train an RL agent in simulation than in reality. 
Simulation to reality (sim-to-real or sim2real) gaps receive much attention recently, in particular, in robotics.
Most RL algorithms are sample intensive
and exploration may cause risky policies to the robot and/or the environment.
However, a simulator usually can not precisely reflect the reality.
How to bridge the sim-to-real gap is critical and challenging.
Sim-to-real is a special type of transfer learning.


Open sources play a critical role in the success of this wave of AI, in particular, RL. 
We list names of some RL benchmarks below, which can be easily found on the Internet:
AI Habitat, Amazon AWS DeepRacer, Deepmind, Behaviour Suite, DeepMind Control Suite, DeepMind Lab, DeepMind Memory Task Suite, DeepMind OpenSpiel, DeepMind Psychlab, Deepmind Real-World Reinforcement Learning, Deepmind RL Unplugged, DeepMind TRFL, Facebook ELF, Google Balloon Learning Environment, Google Dopamine, Google Research Football, Intel RL Coach, Meta-World, Microsoft TextWorld, MineRL, Multiagent emergence environments, OpenAI Gym, OpenAI Gym Retro, Procgen Benchmark, PyBullet Gymperium, Ray/RLlib, RLCard,  RLPy, Screeps, Serpent.AI, Stanford BEHAVIOR, StarCraft II Learning Environment, The Unity Machine Learning Agents Toolkit (ML-Agents), WordCraft.\footnote{See \url{https://neptune.ai/blog/best-benchmarks-for-reinforcement-learning} for a brief discussion for some of the benchmarks.}

In particular, 
Habitat 2.0~\citep{Szot2021Habitat}, 
Isaac Gym~\citep{Makoviychuk2021Isaac}, and 
ThreeDWorld~\citep{Gan2021ThreeDWorld}
provide interactive multi-modal physical simulation platforms.
\cite{Srivastava2021BEHAVIOR} present a benchmark for embodied AI with 100 realistic, diverse and complex household activities in simulation, with the definition, instantiation in a simulator, and evaluation for each activity.
\cite{Ebert2021} propose to boost generalization of robotic skills with cross-task and cross-domain datasets
and evaluate the hypothesis using a bridge dataset with 7,200 demonstrations for 71 tasks in 10 environments.


There are some RL competitions, usually with datasets.\footnote{See e.g., \url{https://github.com/seungjaeryanlee/awesome-rl-competitions}.}
One question remains: Can we have a dataset for RL like the ImageNet for DL?
It is a common wisdom among enterprises and high managements to have big datasets ready to take advantage of the achievements of AI and big data.
It is probably wiser to prepare datasets for reinforcement learning, in particular, for contextual bandits, to avoid losing battles during the highly likely revolutions empowered by RL, the new type of AI.
Different from supervised learning with labelled data, datasets for RL need states/observations, actions and rewards.
Moreover, rewards may arrive later, which causes practical issues~\citep{Agarwal2016}.
The software design needs to consider such issues when preparing datasets for RL.
There are RL competitions in conference like NeurIPS.
NeuIPS 2021 has a new track on Datasets and Benchmarks.

\cite{Lavin2021simulation} present nine motifs of simulation intelligence: 
1) multi-physics and multi-scale modeling, 
2) surrogate modeling and emulation, 
3) simulation-based inference, 
4) causal modeling and inference, 
5) agent-based modeling, 
6) probabilistic programming, 
7) differentiable programming, 
8) open-ended optimization, 
and 9) machine programming. 

Simulation to reality gap, or sim-to-real, or sim2real, or reality gap, attract much attention recently, in particular for robotics.
\cite{Chebotar2019} study how to adapt simulation randomization with real world experience.
\cite{James2019sim2sim} propose to adapt from randomized to canonical scenes, without real-world data.
\cite{James2020RLBench} propose RLBench, a robot learning benchmark and environment.
\cite{Deitke2020RoboTHOR} propose  RoboTHOR, an open sim-to-real embodied AI platform.
\cite{Gondal2019} propose a distanglement dataset to study the sim-to-real transfer of inductive bias.
\cite{Hanna2021} study sim-to-real RL with grounded action transformation.
\cite{Kadian2020Sim2Real} develop a library Habitat-PyRobot Bridge (HaPy) to execute identical code in simulation and on real robots seamlessly, and investigate sim2real predictivity with a new performance metric Sim-vs-Real Correlation Coefficient.
\cite{Zhao2020sim2real} present a brief survey on sim-to-real in deep RL for robotics, 
discussing transfer learning (domain adaptation), domain randomization, knowledge distillation, imitation learning, meta-learning, robust RL. 

Digital twin is related to simulator,
both of which play essential roles in the emerging metaverse~\citep{Lee2021meta}, 
in particular, for the metaverse focusing on science and engineering. 
Metaverse would be the integration of various techniques, including AI, virtual reality (VR), augmented reality (AR), Internet of Things (IoT), cloud computing, communication (e.g., 5G), block chain, etc.
Isaac Gym~\citep{Makoviychuk2021Isaac} is part of Nvidia Omniverse.\footnote{See e.g., \url{https://blogs.nvidia.com/blog/2021/12/14/what-is-a-digital-twin/}}
Habitat 2.0~\citep{Szot2021Habitat}, 
ThreeDWorld~\citep{Gan2021ThreeDWorld},
and BEHAVIOR~\citep{Srivastava2021BEHAVIOR}
can be seen as academic efforts in this directions.


\subsection{Off-policy/Offline Learning}
 
Learning from previously collected data is a feasible approach for some problems,
esp. when there is no perfect model or high-fidelity simulator.
This works to some extent even for challenging problems like AlphaGo~\citep{Silver-AlphaGo-2016}\footnote{As the first work in the AlphaGo series, AlphaGo~\citep{Silver-AlphaGo-2016} initializes the policy with human demonstration data. However, AlphaGo Zero~\citep{Silver-AlphaGo-2017} and AlphaZero~\citep{Silver-AlphaZero-2018} learn from scratch, without human knowledge.} and InstructGPT~\citep{Ouyang2022}.
Moreover, it may be costly, risky, and/or unethical to run a policy in the physical environments, e.g., for healthcare, autonomous driving, and nuclear fusion.
Thus off-policy/offline learning attract significant attention recently.
 
There are two dimensions here: online vs. offline and on-policy vs. off-policy  learning. 
In online learning, an algorithm is trained on data available in sequence. 
In offline learning, or a batch mode, an algorithm is trained on a (fixed) collection of data. 
In on-policy learning, samples are from the same target policy.
In off-policy learning, experience trajectories are usually from some different behaviour policy/policies, but not necessarily from the same target policy. 
For example, both Q-learning and DQN are off-policy learning algorithms.
However, Q-learning is online, while DQN follows a mixture of online mode, by collecting samples on the go, and offline mode, by storing the experience data into a replay buffer and then sampling from it.

In off-policy learning, importance sampling is a standard way to correct the distribution mismatch between the experience data observed from the behavior policy and those unobserved from the target policy. 
\cite{Li2011offline} study the inverse propensity scoring (IPS) with importance sampling in the setting of contextual bandits and apply it to news article recommendation. 
A critical issue for importance sampling is the high variance, esp. for the full RL case~\citep{Precup2000}. 
The doubly robust technique was studied e.g. by \cite{Dudik2011doubly} for contextual bandits, and by \cite{Jiang2016doubly} and \cite{Thomas2016off} for RL, to reduce variance in off-policy learning.
\cite{Liu2018horizon} propose to calculate importance weights on states, rather than on trajectories, to avoid the \enquote{curse of horizon}, i.e., the dependence on the trajectory length.
\cite{Nachum2019DualDICE} study how to estimate discounted stationary distribution ratios, i.e., the likelihood that a certain state action pair appears in the target policy normalized by the likelihood it appears in the dataset. 
\cite{Nachum2020duality} survey duality for RL, including off-policy learning.

There are early examples of offline or batch RL, e.g. least-squares temporal difference (LSTD)~\citep{Bradtke96},  least-squares policy iteration (LSPI) \citep{Lagoudakis03}, and fitted Q iteration~\citep{Ernst2005}. 
These approximate dynamic programming approaches suffer the issue of action distribution shift, and modern offline RL attempts to address it, with
policy constraints~\citep{Kumar2019off},
model-based~\citep{Yu2020off, Kidambi2020off},
value function regularization~\citep{Kumar2020CQL}, and
uncertainty-based methods~\citep{Agarwal2020offline}.
Constraints may be put on distribution, support, and state-action marginal distributions.
Policy constraint methods need to estimate the behavior policy, and tend to be too conservative~\citep{Levine2020offline, Kumar2020offline}.
Model-based offline RL is an alternative approach to avoid unseen outcomes.
See \cite{Levine2020offline} for a survey and \cite{Kumar2020offline} for a tutorial and the reference therein for more details.

\cite{Fujimoto2021} propose to add a behavior cloning term to the policy update of TD3 to regularize the policy, 
aiming to change an online deep RL algorithm minimally to work offline.   
\cite{Ghasemipour2021} propose Expected-Max Q-Learning (EMaQ) for offline and online RL.
\cite{Zhang2021off} propose hyperparameter-free policy selection for offline RL.
\cite{Tang2021off} study model selection for offline RL in healthcare.

The extent of intelligence provided by offline RL is contained by the dataset~\citep{Silver2021}.
A pure offline learning deals with a fixed dataset. 
In practice, we expect an iterative and adaptive way to collect new datasets, at the basis of minutes, hours, or days, etc. and in the extreme, in an online mode, depending on the application. 
\cite{Matsushima2021} propose the concept of deployment efficiency to measure the number of data-collection policies for policy learning, and propose a model-based offline RL approach to achieve deployment-efficiency and sample-efficiency.
\cite{Lee2021offline} propose pessimistic Q-ensemble for offline training and balanced replay for fine-tuning with online data.

\cite{Voloshin2021off} propose the software package Caltech OPE Benchmarking Suite (COBS) with empirical study.
The benchmarking design and methodology include: 
1) design philosophy, w.r.t. design factors of horizon length, reward sparsity, environment stochasticity, unknown behaviour policy, policy and distribution mismatch, and model misspecification;
2) domains, including graph, graph-POMDP, gridworld, pixel-gridworld, mountain car, pixel mountain car, tabular mountain car, and Atari (Enduro);
3) experiment protocol, w.r.t. selection of policies, data generation \& metrics, and implementation \& hyperparameters;  and
4) baselines, including inverse propensity scoring (importance sampling, per-decision importance sampling, weighted importance sampling and per-decision weighted importance sampling), direct methods (model-based estimators, value-based estimators, regression-based estimators, and minimax-style estimators), and hybrid methods.

\cite{Fu2020D4RL} present D4RL, datasets and benchmarks for deep data-driven RL.
\cite{Fu2021DOPE} propose the Deep Off-Policy Evaluation (DOPE) benchmark.
\cite{Gulcehre2020} present RL Unplugged to evaluate and compare offline RL methods.
\cite{Saito2021off} propose Open Bandit Dataset and Pipeline for off-policy evaluation.

Off-policy learning answers what-if questions by counterfactual analysis,  relating to causal inference closely.
\cite{Bareinboim2020} presents a tutorial on causal RL.
\cite{Pearl2018why} introduce cause and effect, or causality.

\subsection{Learning to Learn a.k.a. Meta-learning}

Learning to learn, a.k.a. meta-learning, is learning about some aspects of learning.
It includes concepts as broad as transfer learning, multi-task learning,  one/few/zero-shot learning,
learning to reinforcement learn, learning to optimize, learning combinatorial optimization,
hyper-parameter learning, neural architecture design, automated machine learning (AutoML)/AutoRL/AutoAI, etc.
It is closely related to continual learning and life-long learning~\citep{Khetarpal2020}.
Learning to learn is a core ingredient to achieve strong AI~\citep{Botvinick2019Review, Kaelbling2020, Lake2017, Sutton2019Bitter},
as discussed in Section~\ref{foundation},
and has a long history, 
e.g., \citet{Schmidhuber1987}, \citet{Bengio1991},  and \citet{Thrun1998}.

\citet{Li2017Opt}, along with a blog,\footnote{\url{http://bair.berkeley.edu/blog/2017/09/12/learning-to-optimize-with-rl/}}
divide various learning to learn methods into three categories:
learning what to learn;
learning which model to learn; and
learning how to learn.
The authors mention that, roughly speaking,  learning to learn \enquote{simply means learning something about learning}.
The authors discuss that the term of learning to learn has the root in 
the idea of metacognition by Aristotle in 350 BC,\footnote{\url{http://classics.mit.edu/Aristotle/soul.html}} 
which describes \enquote{the phenomenon that humans not only reason, but also reason about their own process of reasoning}.
In the category of learning what to learn, the aim is to learn  values for base-model parameters, 
gaining the meta-knowledge of commonalities across the family of related tasks, 
and to make the base-learner useful for those tasks~\citep{Thrun1998}.
Examples in this category include methods for transfer learning, multi-task learning, and few-shot learning.
In the category of learning which model to learn,
the aim is to learn which base-model is most suitable for a task,
gaining the meta-knowledge of correlations of performance between various base-models,
by investigating their expressiveness and searchability.
This learns the outcome of learning.
In the category of learning how to learn,
the aim is to learn the process of learning,
gaining the meta-knowledge of commonalities in learning algorithms behaviors.
There are three components for learning how to learn: 
the base-model, the base-algorithm to train the base-model, and the meta-algorithm to learn the base-algorithm.
The goal of learning how to learn is to design the meta-algorithm to learn the base-algorithm, which trains the base-model.
\citet{Bengio1991} and \citet{Li2017Opt} fall into this category.
\citet{Li2017Opt} propose automating unconstrained continuous optimization algorithms with RL.

\citet{Finn2017MAML}, along with a blog,\footnote{\url{https://github.com/cbfinn/maml}}
summarize that there are three categories of methods for learning to learn,
namely, recurrent models, metric learning, and learning optimizers.
In the approach of recurrent models, a recurrent model, e.g., an LSTM, 
is trained to take in data points, e.g., (image, label) pairs for an image classification task, sequentially from the dataset, 
and then processes new data inputs from the task.
The meta-learner usually uses gradient descent to train the learner, 
and the learner uses the recurrent network to process new data.
In the approach of metric learning, a metric space is learned to make learning efficient,
mostly for few-shot classification.
In the approach of learning optimizers, an optimizer is learned, using a meta-learner network to learn to update the learner network to make the learner learn a task effectively.
Motivated by the success of using transfer learning for 
initializing computer vision network weights with the pre-trained ImageNet weights,
\citet{Finn2017MAML} propose model-agnostic meta-learning (MAML)
to optimize an initial representation for a learning algorithm,
so that the parameters can be fine-tuned effectively from a few examples.
MAML works for both supervised learning and RL.

\citet{Duan2017thesis} gives a brief review of meta-learning.
The author discusses meta-learning for supervised learning,
including metric-based models, optimization-based models, and fully generic models,
and other applications.
The author also discusses meta-learning for control,
and proposes to learn RL algorithms and
one-shot imitation learning.

The aim of few-shot meta-learning is to train a model adaptive to a new task quickly,
using only a few data samples and training iterations~\citep{Finn2017MAML}.
Transfer learning is about transferring knowledge learned from different domains, possibly with different feature spaces and/or different data distributions~\citep{Taylor09, Pan2010}.
Curriculum learning~\citep{Bengio2009Curriculum, Narvekar2020, Baker2020HideSeek, Vinyals2019AlphaStar},
model distillation/compression~\citep{Hinton2014, Czarnecki2019}, and sim-to-real  are particular types of transfer learning. 
Multitask learning~\citep{Caruana1997} learns related tasks with a shared representation in parallel, 
leveraging information in related tasks as an inductive bias, to improve generalization, and to help improve learning for all tasks.
\cite{Scholkopf2021} discuss causal representation learning for transfer learning, multitask learning, continual learning, RL, etc.
See \citet{Hutter2019} for a book on AutoML, 
\citet{Hospedales2021} for a survey on meta-learning, 
\citet{Singh2017} for a tutorial about continual learning,
\cite{Chen2021Opt} for a survey and a benchmark on learn to optimize,
and \cite{Portelas2020} for a survey on automatic curriculum learning for deep RL.

\subsection{Explainability and Interpretability}
\label{explain}

Explainability and interpretability are critical for RL/AI for real life.\footnote{Explainability vs. interpretability entails more discussions. As in \cite{Rudin2021},  explainable AI (XAI) \enquote{attempts to explain a black box using an approximation model, derivatives, variable importance measures, or other statistics}, whereas interpretable machine learning creates \enquote{a predictive model that is not a black box}. In \cite{Murdoch2019}, interpretable ML includes explainable ML, intelligible ML, and transparent ML.
\cite{Lipton2018ACM} argues that explanation is post hoc interpretability.
\cite{Miller2019} treats explainability and interpretability as the same.
We refer readers to these reference for more detailed discussions about explainability vs. interpretability, e.g. pages 8-10 in \cite{Rudin2021}.
}
In the following we first summarize several works about interpretable machine learning/AI, then briefly present several papers in interpretable RL.
General principles and challenges from the former are helpful for the latter, and the latter, due to its sequential nature, has its particular characteristics. 

\cite{Miller2019} survey how people define, generate, select, evaluate, and present explanations in philosophy, psychology, and cognitive science and the implication for explainable AI.
The major findings are: 
explanations are contrastive, 
explanation are selected in a biased manner, 
probabilities probably don’t matter,
and explanations are social.
The author summarized that \enquote{explanations are not just the presentation of associations and causes (causal attribution), they are contextual}.

\cite{Doshi-Velez2017} propose to define interpretability \enquote{as the ability to explain or to present in understandable terms to a human}. 
The authors discuss the relationship between interpretability with other desiderata of ML systems.
Fairness or unbiasedness concerns with groups being protected from explicit or implicit discrimination.   
Privacy is about the protection of sensitive information in the data.
An algorithm is reliable and robust if it can achieve a certain level of performance with variation in parameters or inputs.
The predicted change in output due to a perturbation, according to causality, will occur in the real system.
A method is usable if it provides information to help users to accomplish a task.
Trust is about a system with confidence of human users.
Interpretability qualitatively assists to meet these properties: fairness, privacy, reliability, robustness, causality, usability and trust.

\cite{Lipton2018ACM} discusses the desiderata and methods for interpretable AI. 
Desiderata include trust, causality, transferability, informativeness, and fair and ethical decision making.
Techniques and model properties for interpretability include transparency and post hoc explanations.
The different levels of transparency are: simulatability for the entire model, decomposability for individual components such as parameters, and algorithmic transparency for the training algorithm.
Methods for post hoc interpretations include text explanations, visualization, local explanations, and explanation by example.


\cite{Murdoch2019} propose to define interpretable machine learning as \enquote{the extraction of relevant knowledge from a machine-learning model concerning relationships either contained in data or learned by the model}.
The authors propose the predictive, descriptive, relevant framework, with desiderata for evaluation: predictive accuracy, descriptive accuracy, and relevancy judged relative to a human audience. 
The authors propose to categorize existing techniques into model-based and post hoc categories.  
Model-based interpretability constructs models to provide insight into the learned relationship, and includes the following types:
sparsity, simulatability, modularity, domain-based feature engineering and model-based feature engineering.
Post hoc interpretability includes: 1) dataset-level interpretation: 1.1) interaction and feature importances, 1.2) statistical feature importances, 1.3) visualizations, and 1.4) analyzing trends and outliers in predictions, and 2) prediction-Level Interpretation: 2.1) feature importance scores and 2.2) alternatives to feature importances.

\cite{Rudin2021}  propose to define interpretable machine learning in one sentence: \enquote{an interpretable model is constrained, following a domain-specific set of constraints that make reasoning processes understandable}.
The authors discusses principles and challenges of interpretable ML.
In the following we copy from \cite{Rudin2021} without quotation marks.
Principles: 1) An interpretable machine learning model obeys a domain-specific set of constraints to allow it (or its predictions, or the data) to be more easily understood by humans. These constraints can differ dramatically depending on the domain.
2) Despite common rhetoric, interpretable models do not necessarily create or enable trust --- they could also enable distrust. They simply allow users to decide whether to trust them. In other words, they permit a decision of trust, rather than trust itself.
3) It is important not to assume that one needs to make a sacrifice in accuracy in order to gain interpretability. In fact, interpretability often begets accuracy, and not the reverse. Interpretability versus accuracy is, in general, a false dichotomy in machine learning.
4) As part of the full data science process, one should expect both the performance metric and interpretability metric to be iteratively refined.
5) For high stakes decisions, interpretable models should be used if possible, rather than \enquote{explained} black box models.
Ten grand challenges: 1) optimizing sparse logical models such as decision trees; 
2) optimization of scoring systems; 
3) placing constraints into generalized additive models to encourage sparsity and better interpretability; 
4) modern case-based reasoning, including neural networks and matching for causal inference; 
5) complete supervised disentanglement of neural networks; 
6) complete or even partial unsupervised disentanglement of neural networks; 
7) dimensionality reduction for data visualization; 
8) machine learning models that can incorporate physics and other generative or causal constraints; 
9) characterization of the Rashomon set of good models; 
and 10) interpretable reinforcement learning.

In the following we briefly discuss several papers in interpretable RL.
\cite{Amir2019} investigate how to summarize the strategy and expected behaviors in different situations to users, 
with text and/or visualization, proposing a conceptual framework considering intelligent state extraction, world state representation, and strategy summary interface.
\cite{Puri2020} propose to explain moves with saliency maps.
\cite{Atrey2020ICLR} investigate saliency maps, and argue that saliency maps are better for exploration rather than explanation.
\cite{Huber2021} investigate local and global explanations of agent behavior by integrating strategy summaries with saliency maps.
\cite{Dodge2021AAR} propose to assess RL agents by leveraging After-Action Review, a method to assess human agents in many domains.
\cite{Gottesman2019Interpretable} propose approaches to highlight influential transitions.
\cite{Hayes2017Transparency} propose to improve transparency through autonomous policy explanation.
\cite{Huang2019Robots} study how to enable users to anticipate robots' behavior in novel situations.
\cite{Li2019Science} propose a formal methods approach to interpretable RL for robotic planning.
\cite{Madumal2020} propose to generate explanations by learning a structural causal model in model-free RL.
\cite{Mott2019} propose to use attention for interpretability.
\cite{Sequeira2020} study interestingness elements with visual summaries by analyzing four introspection dimensions: frequency, execution certainty, transition-value and sequences.
\cite{Skirzynski2021} discuss automatic discovery of interpretable planning strategies.
\cite{Verma2019ICML} propose to achieve interpretability and verifiability by learning a neural policy first, and then extracting a policy represented by a high-level, domain-specific programming language.
\cite{Heuillet2021} presents a survey on deep RL.

\subsection{Constraints}

Besides predictive and optimal, it is desirable for an AI system to be safe, robust, adaptive, reliable, stable, transparent, fair, trustworthy, explainable,  etc., and not to have behaviours like discrimination w.r.t. race, gender, nationality, etc.
AI for good is a goal.
Some of these properties can be expressed as constraints.
A predictive model is built on domain knowledge, real-world data, and high-fidelity simulators;
a robust method accounts for worst-case scenarios and takes conservative actions,
and an adaptive method learns from online observations and adapts to unknown situations~\citep{Brunke2021Safe}.
See a discussion about the relationship between interpretability with other desiderata of ML systems by \cite{Doshi-Velez2017} in Section~\ref{explain}.

Model-free RL does not assume a prior dynamics model.
The exploration vs. exploitation tradeoff hinders satisfaction of constraints like safety during learning. 
Model-based RL learns the dynamics model, and optimizes a policy.

\cite{Brunke2021Safe} survey safe learning in robotics, from the perspective from learning-based control to safe RL.
Control theory traditionally follows a model-driven approach, leveraging a given dynamics model and providing guarantees w.r.t. known operating conditions. RL traditionally follows a data-driven approach, being adaptable to new contexts at the expense of providing formal guarantees. It is promising to integrate model-driven and data-driven approaches. 
The safe learning control problem is defined by the cost function, the system model, and the constraints, all of which may be initially unknown. 
Data are used to update the control policy or the safety filter.
There are three levels of safety: hard (constraint satisfaction guaranteed), probabilistic (constraint satisfaction with probability), and soft (constraint satisfaction encouraged).
Adaptive control, robust control, and robust Model Predictive Control (MPC)  are control design techniques.
Adaptive control adapts parameters of the controller online to optimize performance, with knowledge of the parametric form of the uncertainties.
Robust control guarantees stability for pre-specified bounded disturbances for dynamics and noise.
Robust MPC extends classical adaptive and robust control by additionally guaranteeing state and input constraints for all possible bounded disturbances.
At every time step, MPC solves a constrained optimization problem and applies the first optimal control decision to the system and then repeats this process in the next time step with the current state.
Tube-based MPC is a common approach for robust MPC, by guaranteeing constraints for true states, which stay inside the tube around nominal states.
RL usually assumes an MDP formulation.
Constrained MDPs and robust MDPs are deployed to satisfy constraints and robustness.


\cite{Brunke2021Safe} reviews the following safe learning control approaches:
a) standard control approaches, 
b) standard RL, 
c) RL encouraging safety and robustness (safe exploration and optimization, risk-averse and uncertainty-aware RL, constrained MDPs and RL, and robust MDPs and RL), 
d) safe model-based RL, 
e) safely learning uncertain dynamics (learning adaptive control, learning robust control, and learning robust MPC), 
f) safety certification (stability and constraint set), 
and g) expressive model with strong safe guarantee (the goal), 
w.r.t. two dimensions:
1) degree of reliance on data (imperfect prior knowledge/model, e.g., dynamics uncertainty), increasing from known dynamics, prior linear dynamics, prior control-affine dynamics, prior structured nonlinear dynamics, prior generic nonlinear dynamics, to unknown dynamics; 
and 2)  degree of safety guarantee, increasing from no guarantees, soft constraint satisfaction, probabilistic constraint satisfaction, to hard constraint satisfaction    

As discussed in \cite{Thomas2019}, to prevent undesirable behaviour of intelligent machines, a user of a standard ML algorithm needs to  constrain the algorithm's behaviour in the objective function (with soft constraints or robust and risk-sensitive methods) or in the feasible set (with hard constraints, chance constraints, or robust optimization methods), both of which requires domain knowledge or extra data analysis. 
\cite{Thomas2019} propose a framework to shift the burden from the user to the designer of the algorithm, by allowing the user to place probabilistic constraints on the solution directly, for classification, regression, and RL.
The author show the performance of the proposed method with experiments of grade point averages (GPAs) prediction for classification and bolus calculation in type 1 diabetes for RL.

\cite{Wiens2019} discuss how to do no harm in the context of healthcare, which may somewhat generalize to AI.
RL/AI practitioners, esp. those with AI power and resources and/or those dealing with high-stake issues like healthcare and autonomous driving, may need to take a \enquote{Hippocratic oath} or even go under stricter regulation.

\cite{Garcia2015safety} present a survey on safe RL.
\cite{Wing2021trust} reviews trustworthy AI. 
\cite{Russell2019} discusses human compatible AI.
\cite{Mitchell2020} present a guide for thinking humans for AI.
\cite{Altman1999} is a book about constrained MDP. 
\cite{Szepesvari2020constraints} discusses multi-objective and constrained RL.
 

\subsection{Software Development and Deployment}
\label{software}

AI and IT software systems are (very) different.
A piece of IT code can call APIs to achieve a certain goal, while, at least at the current stage, AI coding still needs considerable manual fine-tuning, for data, features, hyper-parameters, training methods, etc.      
There are already discussions about software and systems issues for machine learning systems. 
However, there are not many papers discussing systems issues for RL, with e.g.,~\cite{Agarwal2016} (discussed in Section~\ref{bandits}) and RLlib as exceptions, 
which call for more research and development (R\&D). 
One factor may be that ML systems are much more widely deployed than RL systems.

Ray/RLlib~\citep{Liang2018RLlib} is an RL library to achieve a composable hierarchy for distributed programming, aims for a production level.
RLlib Flow~\citep{Liang2021RLlib} treats distributed RL as a dataflow and targets for more efficient programming.

\citet{Henderson2018} investigate reproducibility, experimental techniques, and reporting procedures for deep RL.
 \cite{Engstrom2020PPO} study the importance of code-level optimizations in deep policy gradient algorithms, in particular, PPO and TRPO.
 \cite{Ilyas2020PG} observe mismatch between predicted and empirical behavior in deep policy gradient algorithms w.r.t. gradient estimation, value prediction, and optimization landscapes.
\cite{Andrychowicz2021AC} investigate lower level design decisions for on-policy deep actor-critic implementation to understand their impact on agent performance.
\cite{Huang2022PPO} discuss many implementation details of PPO.
These works suggest that it is essential to have a deeper understanding of deep RL methods beyond evaluation with benchmarks in developing an RL toolkit.

\cite{Sculley2014} discuss the hidden technical debt in machine learning systems.
A technical debt refers to the long-term hidden costs accumulated from expedient yet suboptimal decisions in the short term.
Besides normal code complexity issues in traditional software systems, 
ML systems incur at the system level enormous technical debts, 
e.g., boundary erosion,
entanglement,
hidden feedback loops,
undeclared consumers,
data dependencies, 
configuration issues,
changes in the external world, and
system-level anti-patterns.
Potential solutions include
refactoring code,
improving unit tests,
deleting dead code,
reducing dependencies,
tightening APIs, and
improving documentation. 
In Section~\ref{bandits}, we discuss Decision Service, which aims to achieve low technical debts with contextual bandits.  

\cite{Amershi2019ICSE} discuss software engineering for machine learning.
The authors illustrate nine stages of the machine learning workflow, namely, model requirements, data collection, data cleaning, data labeling, feature engineering, model training, model evaluation, model deployment, and model monitoring.
There are many feedback loops in the workflow, e.g., between model training and feature engineering,
and model evaluation and model monitoring may loop back to any previous stages.
The authors also identify three major differences in software engineering for AI from that for previous software application domains:
1) it is more complex and more difficult to discover, manage, and version data;
2) different skill set for model customization and model reuse; and
3) AI components are less modular and more entangled in complex ways.

\cite{Ng020Gap} discusses how to bridge AI’s proof of concept to production gap w.r.t. small data, generalization and change management.
\cite{Ng020Gap} presents the ways to manage the change the technology brings: budget enough time, identify all stakeholders, provide reassurance, explain what’s happening and why, and right-size the first project, and argue that key technical tools are explainable AI and auditing.
\cite{Ng020Gap} discusses that production AI projects require more than ML code,
e.g., data verification, environment change detection, data and model version control, process management tools and detect model performance degradation.

\cite{Ng2021MLOps} argues that AI system = Code + Data and MLOps' most important task is to make high quality data available through all stages of the ML project lifecycle, including scoping (decide on problem to solve), data (acquire data for model), modelling (build/train AI model), and deployment (run in production to create value).
\cite{Ng2021MLOps} discusses that model-centric AI focus on the question: \enquote{How can you change the model (code) to improve performance?},
while data-centric AI focus on the question: \enquote{How can you systematically change your data (input x or labels y) to improve performance?}.
\cite{Ng2021MLOps} argues that an important frontier is to build MLOps tools to make the process of data-centric AI efficient and systematic.

\cite{Sambasivan2021} discuss data cascades, i.e., compounding events from data issues cause negative, downstream effects, leading to technical debts over time.
\cite{Paleyes2020} discuss deployment challenges for machine learning, w.r.t.
data management, model learning, model verification, model deployment, and cross-cutting aspects.
\cite{Zhang2022testing} discuss machine learning testing.
MLSys is a conference dedicated to ML systems.\footnote{\url{https://mlsys.org}}

\subsection{Business Perspectives}

In this subsection, we discuss business perspectives of RL and AI in general.

\subsubsection*{RL is Promising!}

In a Mckinsey Analytics report titled \enquote{It’s time for businesses to chart a course for reinforcement learning}~\citep{Corbo2021}, the authors start with a story that RL helped reigning sailing Emirates Team New Zealand secure the fourth champion for 2021 America’s Cup Match.
The authors then argue that RL will potentially deliver value in every business domain and industry; in the near-term, RL is particularly helpful for decision makings in design and product development, complex operations, and customer interactions;
and for wide-scale adoption, we need a good learning algorithm with a reward function, a learning environment like a simulator, a digital twin or a digital platform, and compute power.
RL becomes more accessible as technological advances w.r.t. algorithm efficiency, compute cost, and support from cloud computing.
The authors give advices to leaders about how to start with RL: 
1) find the right problem, solve the challenges not addressed by traditional methods;
2) consider compute costs upfront;
3) future-proof the simulator or digital twin for RL;
4) human in the loop, to improve humans' performance, and to leverage experts' domain knowledge;
5) identify and manage potential risks, such as explainability and ethical concerns.  

In a Harvard Business Review article, \cite{Hume2021} predict that RL is the next big thing, 
list tough problems companies are addressing, e.g., 
trade execution platform for multiple strategies, test schedules for business partner devices, recommendation engine, financial derivative risk and pricing calculations, data center cooling, and order dispatching,  
and suggest the way to spot an opportunity for RL:
make a list,
consider other options,
be careful what you wish for,
ask whether it’s worth it, and
prepare to be patient.

In an MIT Technology Review article, \cite{Heaven2021} discusses that in supply chains, simulations, digital twins, together with AI, in particular, deep RL, can help improve visibility, balance efficiency and resiliency, conduct counterfactual or what-if analysis, and undergo stress tests, to help mitigate the negative effect of disruptions, in particular, during the pandemic. 
The article mentions that David Simchi-Levi said \enquote{a million dollars and 18 months can give you many of the benefits}.

\subsubsection*{There are Still Business Issues for ML/AI Systems Though.}

In an Andreessen Horowitz blog, \cite{Casado2020} argue that AI creates a new type of business,
combining both software and services, with low gross margins, scaling challenges, weak defensive moats,
caused by heavy cloud infrastructure usage, 
ongoing human support,
issues with edge cases,
the commoditization of AI models,
and challenges with data network effects.
The author presents practical advices for founders:
1) eliminate model complexity as much as possible,
2) choose problem domains carefully to reduce data complexity,
3) plan for high variable costs,
4) embrace services plan for change in the tech stack,
and 5) build defensibility the old-fashioned way.

\subsubsection*{AI$\infty$, AI$x$, and AIZero}

For the current practice in AI, we make the following categorization: AI$\infty$, AI$x$, and AIZero.
Many deep learning, or big data methods, like AlexNet, relies on huge amount of labelled data.
Model-free RL interacts with the environment online or offline to collect a huge amount of training data.
We classify them as AI$\infty$, which requires significant manual efforts or interactions with the physical system to generate training data.
When there is a perfect model, we can build a perfect simulator to generate training data, in a digital or virtual world, which is much less costly than in a physical world.
AlphaGo Zero/AlphaZero and those dynamic systems with perfect partial differential equations are in this category.
We classify them as AIZero, i.e., once the data generator is set up, it can generate data without manual effort, and without the cost of interacting with the physical system.
Model-free RL for Atari games, e.g., with ALE, appears as AI$\infty$ for the agent.
However, for the whole system, we can say it is AIZero, since we have a perfect simulator.
There are ways to help improve data efficiency, e.g., simulation, digital twins, self-supervised learning,  and for RL in particular, intrinsic motivation, auxiliary signals, and model-based methods.
We call such approaches as AI$x$, where $x$ is a number between 0 and $\infty$, indicating the degree of inaccuracy of the model involved, or the amount of effort of \enquote{manual} collecting of data, or the amount of effort of interacting with physical systems.
Admittedly, $x$ is only loosely defined.
AI$x$ approaches AIZero as the underlying models approach perfect.
GPT-3 is in AI$x$. 
Since InstructGPT involves human feedback, it is a combination of AI$\infty$ and AI$x$. 
AlphaGo involves human demonstration data and a perfect model/simulator, so it is a combination of AI$\infty$ and AIZero.

Considering the Gartner Hype Cycle, arguably, AlexNet in 2012 was in the stage of Technology Trigger. 
Shortly after AlphaGo, it reached Peak of Inflated Expectation, with bubbles from all sorts of fundings.
It may be in Trough of Disillusionment now.
We are witnessing that deep learning, RL and AI in general are making steady progress,
as more and more scenarios moving from AI$\infty$ to AI$x$ and some approaching AIZero.
Consequently, AI gradually goes into Slope of Enlightenment and Plateau of Productivity.


\subsection{More Challenges}


As illustrated in Figure~\ref{issues}, there are many more important topics, like sample efficiency, credit assignment,  
which are discussed explicitly or implicitly in other sub-sections, as well as collective intelligence~\citep{Millhouse2021}, evolution~\citep{Salimans2017evolution, Lehman2020, Ridley2015}, etc.
See \cite{Levine2021} for a discussion about real world RL.
We discuss a few topics briefly in the following.

\subsubsection*{Generalization}

Generalization in RL is nascent yet vital.
\cite{Kirk2021}  is a recent survey, with the focus on zero-shot policy transfer.
A benchmark includes an environment and an evaluation protocol.
There are three types of environments: 1) singleton, where training and testing environments are identical; 
2) independent and identically distributed (IID), where training and testing distributions are identical; 
and 3) out-of-distribution (OOD), where training and testing distributions are different.
The authors present a taxonomy of benchmarks for generalization in RL, 
w.r.t. style (2D, 3D, Arcade, continuous control, driving, grid, language-conditioned, LQR, structured, or text), 
context (continuous, discrete cardinal, discrete ordinal, or procedural content generation (PCG)), 
and variation (state, dynamics, observation, or reward function).
An evaluation protocol specifies the training and testing contexts and sampling from the training context set,
for PCG or controllable environments.
The authors recommend to combine PCG and controllable factors of variation.
The authors discuss methods for generalization in RL:
1) increasing similarity between training and testing: 1.1) data augmentation and domain randomization, 1.2) environment generation, and 1.3) optimization objectives; 
2) handling differences between training and testing: 2.1) encoding inductive biases, 2.2) regularization and simplicity, 2.3) learning invariances, and 2.4) adapting online; 
and 3) RL-specific problems and improvements: 3.1) RL-specific problems and 3.2) better optimization without overfitting.
The authors introduce the concepts of context efficiency, about the number of context,  analogous to sample efficiency, about number of samples during training. 
The authors list the following future directions for generalization RL:
fast online adaptation, tackling RL-specific issues, novel architectures, model-based RL, environment generation, offline RL, reward-function variation, context-efficiency, and continual RL. 


\subsubsection*{Multi-agent RL}

Multi-agent RL (MARL) is the integration of multi-agent systems~\citep{Shoham2007, Hernandez-Leal2019, , Zhang2021MARL} with RL. 
It is at the intersection of game theory~\citep{Leyton-Brown2008} and RL/AI.  
Multi-agent systems are a great tool to model interactions among agents,
for competition, cooperation and a mixture of them,
with rich applications in human society.
Besides issues in RL like sparse rewards and sample efficiency, there are new issues like multiple equilibria, and even fundamental issues like what is the question for multi-agent learning, and whether convergence to an equilibrium is an appropriate goal, etc. 
Consequently, multi-agent learning is challenging both technically and conceptually, and demands clear understanding of the problem to be solved, the criteria for evaluation, and coherent research agendas~\citep{Shoham2007}.
\cite{Hernandez-Leal2019, Zhang2021MARL} are recent surveys about multi-agent RL.
\cite{Papoudakis2019} survey non-stationarity in multi-agent deep RL.


\subsubsection*{Hierarchical RL}

Hierarchical RL~\citep{Pateria2022} is a way to learn, plan, and represent knowledge with temporal abstraction at multiple levels, with a long history, e.g., options~\citep{Sutton1999}. Hierarchical RL is an approach for issues of sparse rewards and/or long horizons, with exploration in the space of high-level goals. The modular structure of hierarchical RL approaches is usually conducive to transfer and multi-task learning. The concepts of sub-goal, option, skill, and, macro-action are related. Hierarchical planning is a classical topic in AI~\citep{Russell2009}.

\subsubsection*{Neuro-symbolic AI}

The major achievements in AI in the last ten years or so, in particular, with deep learning, are criticized for lacking of logic and reasoning.
Most deep learning approaches are finding correlations or associations, rather than causality or interventions and counterfactuals.
Neuro-symbolic approaches are promising to encode logic in learning mechanisms. 
\cite{Garcez2020}  propose that a neurosymbolic approach is critical for the next wave of AI. 
\cite{Hochreiter2022} introduce the concept of \enquote{broad AI}, in contrast to and between \enquote{narrow AI} and \enquote{general AI}, and highlight the importance of neuro-symbolic AI, in particular, with graph neural networks~\citep{Lamb2020GNN}.
In the following, we directly quote from two papers.
\enquote{The burgeoning area of neurosymbolic AI, which unites classical symbolic approaches to AI with the more data-driven neural approaches, may be where the most progress towards the AI dream is seen over the next decade.}~\citep{Littman2021}
\enquote{How are the directions suggested by these open questions related to the symbolic AI research program from the 20th century? Clearly, this symbolic AI program aimed at achieving system 2 abilities, such as reasoning, being able to factorize knowledge into pieces which can easily recombined in a sequence of computational steps, and being able to manipulate abstract variables, types, and instances. We would like to design neural networks which can do all these things while working with real-valued vectors so as to pre-serve the strengths of deep learning which include efficient large-scale learning using differentiable computation and gradient-based adaptation, grounding of high-level concepts in low-level perception and action, handling uncertain data, and using distributed representations.}~\citep{Bengio2021}

\subsubsection*{How to select RL algorithms?}

Before ending, let's pose a question: How to select RL algorithms?

There are more and more interests in RL from academia, industry, governments, and venture capitals (VC). 
The question about RL algorithm selection 
may be from a student wanting to try a toy example/benchmark,
or from a company planning to set up a prototype to see how RL works,
or from a government working on an AI project like smart city,
or from entrepreneurs thinking to build a startup leveraging the potential of RL.

There are so many algorithms, w.r.t. so many dimensions, like on/off policy, model-free/based, online/offline, value function/policy optimization, exploration methods, etc.
There may also be considerations for representation, end-to-end or not, auxiliary or un/self-supervised reward, prior knowledge, meta-learning, hierarchical, multi-agent,  causal, symbolic, relational, explainable,  safety, robustness, AI for good, etc. 

How about algorithms of the same/similar type, e.g., how to select among A3C, DDPG, PPO, SAC, TD3? Rather than trying everything, are there any refined suggestion?

Some factors may not be hard to characterize to select some type(s) of algorithms for certain problems. Some may be non-trivial; maybe being at the frontier of research actually.
Some factors may be critical for making RL practical, like sample efficiency, safety, and explainability. 

 \cite{Badia2020} and \cite{Laroche2018} use bandits to choose parameterized algorithms. 
Benchmarking/comparison papers may be helpful, e.g., \cite{Duan2016}, \cite{Mahmood2018}, \cite{Colas2019} and \cite{Agarwal2021}.
As in a survey about automated RL (AutoRL)~\citep{ParkerHolder2022},
we may automate task design, algorithms, architectures, and hyperparameters,
and AutoRL methods include the following: 
random/grid search, Bayesian optimization, evolutionary approaches, population based training, meta-gradients, blackbox online tuning, learning algorithms, and environment design. 


An AutoRL approach may not be fully satisfactory.
It is desirable to have general, intuitive, maybe \enquote{rule-based} or \enquote{explainable}, guidelines for selecting RL algorithms.
Or we are not there yet, i.e., we still need to design better algorithms.
In Section~\ref{software}, we discuss implementation issues~\citep{Henderson2018, Engstrom2020PPO, Ilyas2020PG, Andrychowicz2021AC, Huang2022PPO}. 
The right question may be: How to \emph{design} RL algorithms?

\section{Discussions}
\label{discussion}

 There are stunning news for reinforcement learning like AlphaGo and some successful applications in production.
There are growing interests in research in RL for real life with exciting achievements,
attempting to address challenges as we discussed.
However, there is still the question:  
Why has RL not been widely adopted in practice yet?
Here is an attempt to answer it. 
And we hope to ignite more discussions.

One key issue is, RL does not have an \enquote{AlphaGo moment} in practice yet, or does not have a killer application, a metaphor for significant business value, which drives the wide adoption of a technology, esp. in a free market.  
Deep learning had the \enquote{AlexNet moment}, with a significant performance improvement for the ImageNet, and with image recognition as the killer application.
AlphaGo was a headline news worldwide.
However, it appears that it does not have direct, significant business value.
We have witnessed various successes applying RL in practice.
However, there is not something for RL equivalent to face or speech recognition for deep learning yet. 
The market still needs to validate if RL can add significant business value, esp. comparing with techniques in traditional machine learning/AI, optimal control or operations research.
For example, for recommender systems, supervised and unsupervised machine learning and data mining techniques are competitors.
For problems like robotics, transportation and supply chain, reinforcement learning needs to outperform optimal control and operations research (significantly) to be (widely) adopted.
Will the nuclear fusion be a killer application for RL? 
Likely not, since it requires huge resources, harder to access than thousands of GPUs.
However the success definitely shows the power of RL, and will attract more attention to RL, in particular, in the optimal control community. 
This being said, there is no doubt that RL has great value, which underlies its recent burgeoning progress, esp. in academia.

A fundamental issue is that there are still challenges with theories, algorithms and implementations.
This is also the issue for deep learning, in some sense; however, empirical performance in wide applications justifies the wide adoption of deep learning.
Off-the-shelf RL algorithms may not apply to real life problems directly, and still need significant fine-tunings from RL experts.
We still encounter technical debts in software engineering and system deployment  for RL,
which is still nascent for research and development (R\&D), 
partly as a consequence of not having many deployed RL systems yet.
RL calls for better theories and algorithms as always, and more and deeper integration with real life applications and more deployed systems.

The R\&D of RL projects requires considerable resources, including talents, compute, and funding, which are still insufficient. 
RL calls for more investments, which still need long-term thinking and the spirit of trial and error.

It is normal for a new technology to be adopted slowly, esp. by traditional industrial sectors. 
Should the technical route for RL be RL+ or +RL? 
With RL+, an RL expert leads the project, involving domain experts; 
whereas with +RL, RL experts need to collaborate with domain experts.
Ideally experts have deep understanding of both RL/AI and domain knowledge.
It is relatively easy to have such versatile experts for areas like recommender systems and robotics, since many AI people are working in these areas.
In such cases, RL+ is possible.
However, for areas without many AI experts yet, it may be more practical for domain experts to have some knowledge of RL/AI, 
e.g., with the capability of managing an RL/AI application project,
and coordinating RL/AI experts to help solve the RL/AI problem abstracted from the problem in hand, 
providing technical details of domain knowledge to RL/AI experts.
This applies particularly to domains not easy for an RL/AI expert without sufficient background to have a decent understanding quickly, e.g., drug design, retrosynthesis, genetics, or arts.
In such cases, likely it is +RL. 
RL calls for more interdisciplinary education, training, and collaborations.

An AI project, esp. an RL one, is in stark contrast with an IT project, in that
it is desirable for all people involved,  from engineer to CEO, to have some knowledge of RL/AI. 
This is natural for technical people.
For management people, with decent knowledge of RL/AI, it is easy for them to appreciate the oppotunities and understand the challenges of the technology, and it is convenient for them to communicate with technical people.
One issue is, the learning curve for RL is much steeper than for deep learning, for both concepts and technical details.
RL calls for more education and training, esp. for non-technical people.

Similar to AI, an RL business may also be a combination of software and service, with low gross margins, scaling challenges,  and weak defensive moats. 
RL calls for understanding of, respect for, and if possible, innovations in, the business model. 

From the discussion above, we see a big chicken-egg problem here: no killer application, insufficient appreciation from higher management, insufficient resources, insufficient R\&D, slow adoption, then, as a result, non-trivial to have a killer application. There are underlying challenges from education, research, development and business.

\begin{figure}[h]
\includegraphics[width=1.0\linewidth]{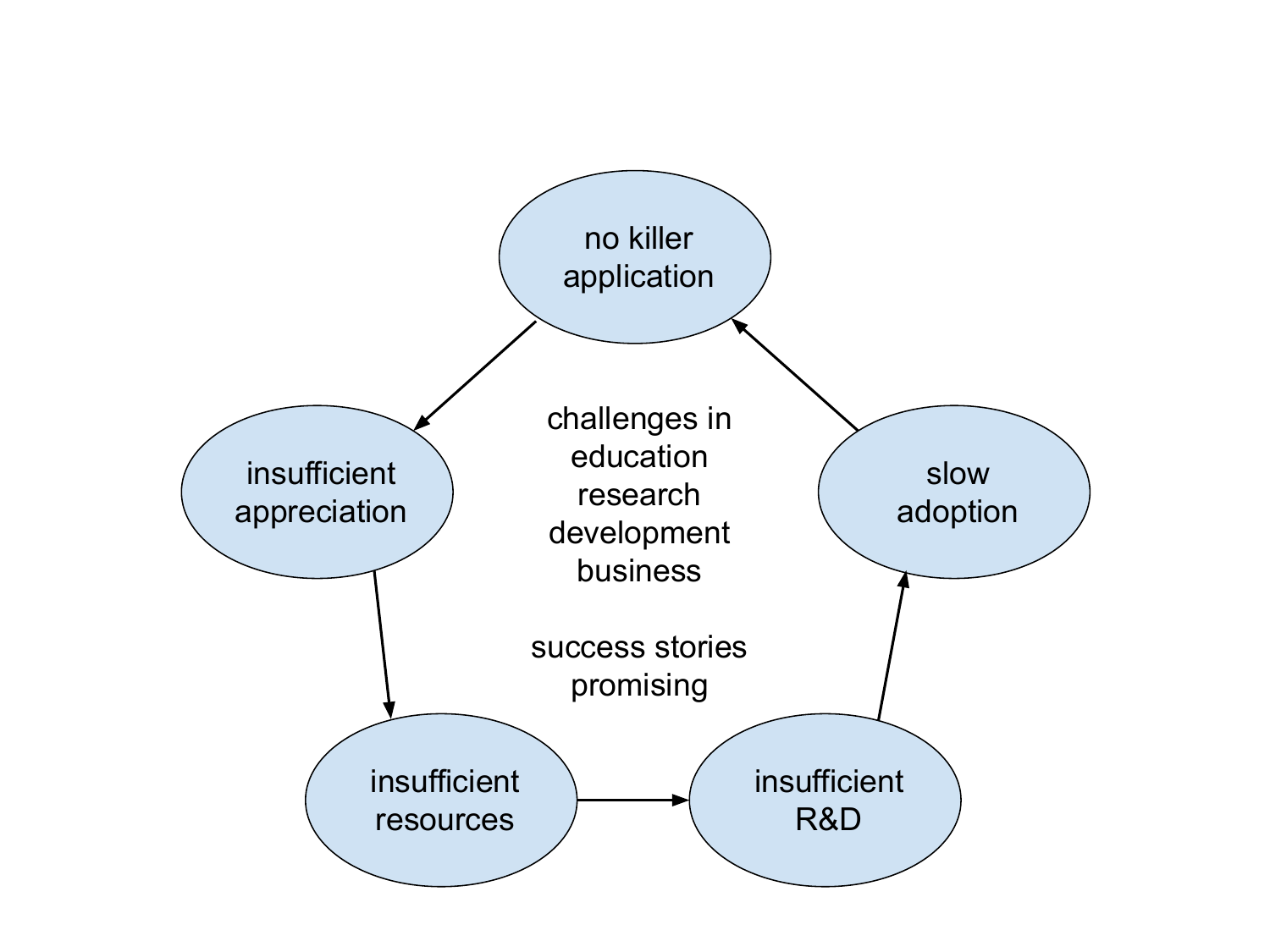}
\caption{Why has RL not been widely adopted in practice yet? The big chicken-egg loop: no killer application, insufficient appreciation from higher management, insufficient resources, insufficient R\&D, slow adoption, then, as a result, non-trivial to have a killer application. There are underlying challenges from education, research, development and business. The status quo is actually much better, having many success stories. With challenges, RL is promising.}
\end{figure}

The status quo is actually much better than this negative feedback loop, thanks to those people and organizations who are forward looking and venture their belief in the potential of reinforcement learning. 
Besides funding agencies, there are companies making big investments in RL, like Deepmind, Google, Microsoft, Facebook,  OpenAI, Nvidia, Didi,  just name a few.  
There are also a few startups working on RL. 
More and more researchers, engineers and managers are interested in RL for real life.
And remember that, the killer applications of face/speech recognition for deep learning was born out of the last AI winter, in particular, for (deep) neural networks.
It is possible that there may not be an AlphaGo moment in practice or an RL killer application, and RL may be permeating gradually yet widely.
It is also possible that RL may not be the major technique, but plays a complementary role, for some period of time, esp. in the whole landscape of an AI business.
We are positive, although reinforcement learning is not quite there yet for the prevalence of real life applications.

Before closing, we attempt to answer the question: When is RL helpful?

In general, RL can be helpful, if a problem can be regarded as or transformed to a sequential decision making  problem, and states/observations, actions, and (sometimes) rewards can be constructed.
RL is helpful for goal-directed learning.
RL may help to automate and optimize previously manually designed strategies.
The rate of success expects to be higher if a problem comes with a perfect model,  a high-fidelity simulator, or a large amount of data. 

For the current state of the art, RL is helpful when big data are available, from a model, from a good simulator/digital twin, or from interactions (online or offline).
For a problem in natural science and engineering,
the objective function may be clear, with a standard answer, and straightforward to evaluate.
For example, AlphaGo has a perfect simulator and the objective is clear, i.e., to win.
Good models/simulators usually come with many problems in combinatorial optimization, operations research, optimal control, drug design, etc.
(Admittedly, there may be multiple objectives or reward is non-trivial to specify.)
In contrast, for a problem in social science and arts,
usually there are people involved, thus the problem is influenced by psychology, behavioural science, etc., 
the objective is usually subjective, may not have a standard answer, and may not be easy to evaluate.
Example applications include education and game design. 
Recommender systems usually have bountiful data, so there are successful production systems.
However, there are still challenges for those facing human users.
Recent progress in RL from human feedback has shown its benefits in human value alignment.
Concepts like psychology, e.g. intrinsic motivation, flow and self-determination theory, may serve as a bridge connecting RL/AI with social science and arts, e.g., by defining a reward function.

With challenges, we see great opportunities ahead for reinforcement learning.
From Dimitri Bertsekas, \enquote{We can begin to address practical problems of unimaginable difficulty!} and \enquote{There is an exciting journey ahead!}

\section*{Acknowledgement}

This article benefits from previous study, research, discussions, talks, conferences, workshops, etc.
The author thanks several group of people: 1) the invited speakers, panelists, authors, reviewers, attendants and co-organizers of the Reinforcement Learning for Real Life (RL4RealLife) workshops; 
2) authors, reviewers and guest co-editors of special issues of the Machine Learning journal; 
3) organizers and audience of talks the author gave in the last several years; 
and 4) all researchers and practitioners in reinforcement learning, esp. in RL4RealLife; 
in particular, 
Alekh Agarwal, 
Ofra Amir, 
Craig Boutilier,
Matthew Botvinick, 
Michael Bowling,
Emma Brunskill,
Craig Buhr, 
Hasham Burhani,
Michael Buro,
Tianyou Chai,
Ed Chi, 
Minmin Chen,
Jim Dai, 
Thomas Dietterich, 
Maria Dimakopoulou,
Finale Doshi-Velez,
Gabriel Dulac-Arnold,
Fei Fang, 
Hai Fang,
Alan Fern, 
Chelsea Finn,
Zhanghua Fu, 
Jason Gauci,
Alborz Geramifard,
Mohammad Ghavamzadeh,
Omer Gottesman,
Haoji Hu,
Ruitong Huang,
Zhimin Hou,
Leslie Pack Kaelbling,
George Konidaris, 
John Langford, 
Lihong Li,
Xiaolin (Andy) Li,
Xin Liu,
Yao Liu,
Zachary C. Lipton, 
Zongqing Lu, 
Rupam Mahmood,
Shie Mannor,
Jeff Mendenhall,
Martin M{\"u}ller,
Susan Murphy,
Anusha Nagabandi, 
Xubin Ping,
Warren Powell,
Niranjani Prasad, 
Zhiwei (Tony) Qin,
Martin Riedmiller,
Suchi Saria,
Angela Schoellig, 
Dale Schuurmans,
David Silver,
Peter Stone,
Richard Sutton,
Csaba Szepesv{\'a}ri, 
Matthew E. Taylor,
Georgios Theocharous,
Yuandong Tian, 
Ivor Tsang,
Tao Wang,
Huan Yu,
Rose Yu, 
Yang Yu, 
Xiangyu Zhao,
and Hongtu Zhu.
Any errors are my own.
Comments and criticisms are always welcome.


\bibliography{RLbib}

\bibliographystyle{apa}

\end{document}